\documentclass{article} 
\usepackage{iclr2027_conference,times}

\usepackage{amsmath,amsfonts,bm}

\def\eqref#1{equation~\ref{#1}}

\def\1{\bm{1}}

\DeclareMathAlphabet{\mathsfit}{\encodingdefault}{\sfdefault}{m}{sl}
\SetMathAlphabet{\mathsfit}{bold}{\encodingdefault}{\sfdefault}{bx}{n}

\PassOptionsToPackage{hyphens}{url}

\usepackage{hyperref}
\usepackage{url}

\usepackage{amsmath}
\usepackage{amssymb}
\usepackage{graphicx}
\usepackage{booktabs}
\usepackage{multirow}
\usepackage{array}

\usepackage{wrapfig}

\newcommand{\dH}{\ensuremath{\Delta\mathrm{H}}}
\newcommand{\dC}{\ensuremath{\Delta\mathrm{C}}}
\newcommand{\dQ}{\ensuremath{\Delta\forall}}
\newcommand{\dT}{\ensuremath{\Delta\mathrm{T}}}
\newcommand{\dR}{\ensuremath{\Delta\mathrm{R}}}
\newcommand{\dI}{\ensuremath{\Delta\mathrm{I}}}
\newcommand{\Cauto}{\ensuremath{C_{\mathrm{auto}}}}
\newcommand{\Set}{\ensuremath{\mathbf{Set}}}
\newcommand{\Mone}{\textsc{m1}}
\newcommand{\Mtwo}{\textsc{m2}}
\newcommand{\DSI}{\textsc{dsi}}

\title{Objects Without Morphisms: What LLMs for Mathematics Do Not Represent}

\author{Yanli Wang \\
Imperial College London \\
\texttt{yanli.wang22@imperial.ac.uk}
\And 
Suijin Wang \\
Tsinghua University \\
\texttt{wangsj24@mails.tsinghua.edu.cn}
\And 
Xiaopeng Yuan \\
University of Illinois at Urbana-Champaign \\
\texttt{xyuan30@illinois.edu}
\And 
Haohan Wang \\
University of Illinois at Urbana-Champaign \\
\texttt{haohanw@illinois.edu}
}

\iclrfinalcopy 
\begin{document}
\maketitle
\lhead{Preprint}

\begin{abstract}
Large language models (LLMs) have reached expert-level performance on competition mathematics largely through the volume of search placed around them: candidate solutions are sampled in quantity and retained only when an external criterion accepts them. Such a procedure improves the outcome that survives it while leaving untouched what the model represents. We examine that question where no external criterion exists: translating statements between the dialects of neighbouring subfields, where fidelity turns on the \emph{level of generality} at which content is asserted. The source leaves that level implicit in its vocabulary, so a faithful translation must recover it from the relation between the theories. We introduce an instrument that codes truth, content and scope in separate blind queues, with a judge-free measure of whether a rewrite states the hypothesis implicit in its source, and establish its sensitivity with a planted-positive control. Across seven models from four families, translating towards the general framing widens the domain of quantification in $60.6\%$ of rewrites and narrows it in none; translating towards the concrete framing narrows it in $28.3\%$ and widens it in $0.3\%$. The hypothesis that would prevent it is stated in $21.6\%$ of model rewrites and $4.2\%$ of human statements. Capability does not govern the asymmetry: it appears in every model tested, and the most capable widens least. It replicates on the half of the benchmark held out by a pre-registered rule, and on statements written by mathematicians. Instructing a model to state every hypothesis it requires raises that rate but not its sensitivity to direction. We argue that these systems have acquired an object-level correspondence between subfield vocabularies without the constraint under which a translation between theories carries hypotheses to hypotheses.
\end{abstract}

\section{Introduction}
\label{sec:intro}
\begin{figure}[t]
\begin{center}
\includegraphics[width=0.74\textwidth,trim=44 42 44 40,clip]{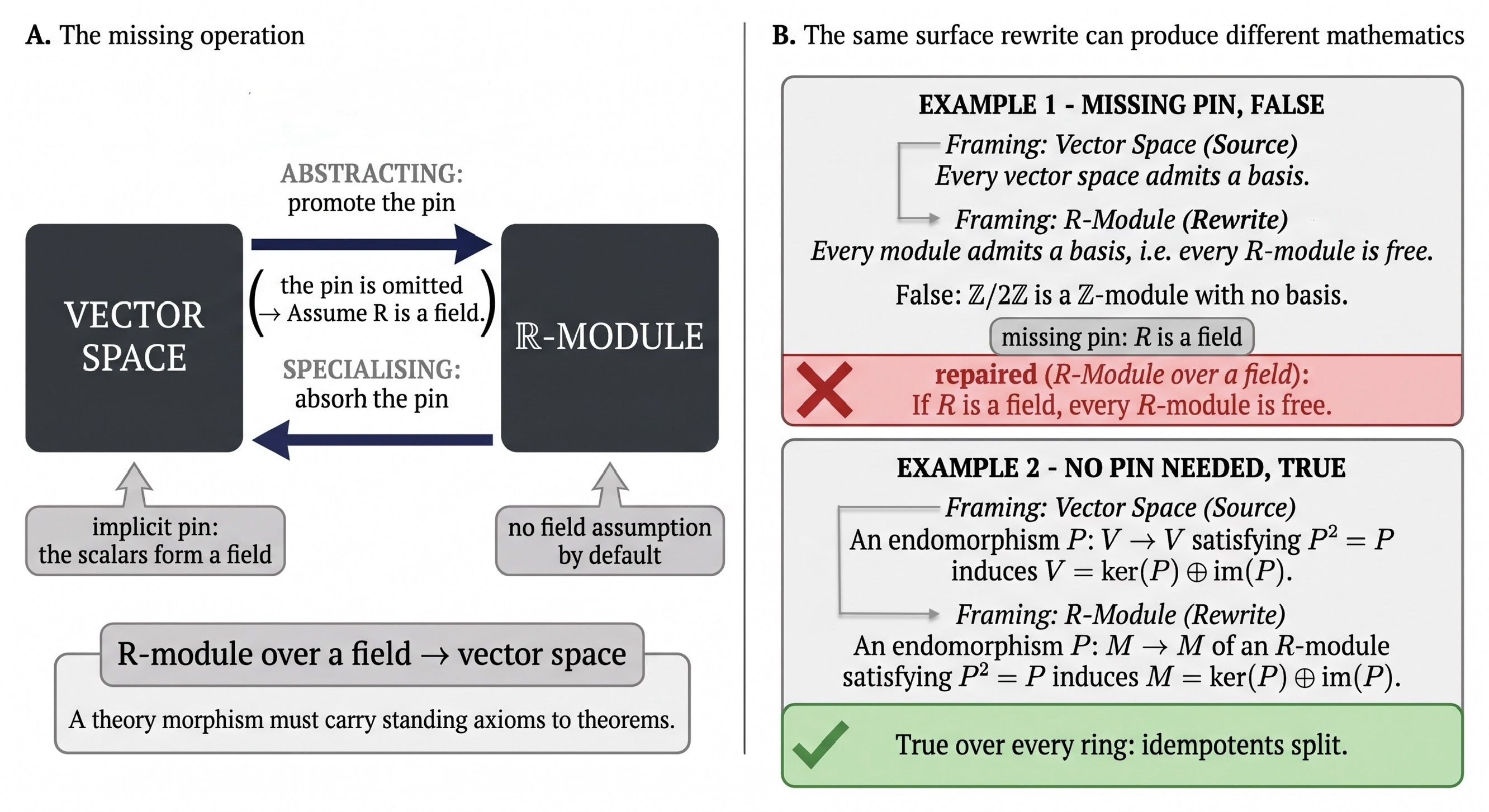}
\end{center}
\caption{\textbf{The operation a cross-framing rewrite must perform, and why its failure escapes single-scale evaluation.} \textbf{(A)} ``Vector space'' \emph{means} ``module over a field'': the field hypothesis is encoded in the noun, so a rewrite into module theory must \textbf{promote} it to an explicit statement, and the reverse rewrite must \textbf{absorb} it into the vocabulary. We call this hypothesis the \emph{pin}. \textbf{(B)} The \emph{same} surface rewrite, $V\mapsto M$, applied to two corpus items. Above, the pin is not promoted and the result is false, since $\mathbb{Z}/2\mathbb{Z}$ is a $\mathbb{Z}$-module without a basis (the repair is shown in red). Below, no pin is required, because idempotents split over every ring, and the identical transformation yields a theorem. The outcome is determined by the mathematics rather than by anything the model varied.}
\label{fig:mathstructure}
\vspace{-5mm}
\end{figure}

Mathematical reasoning is among the most extensively studied capabilities of large language models. Benchmarks of grade-school and competition problems \citep{cobbe2021trainingverifierssolvemath, hendrycks2021measuringmathematicalproblemsolving} were saturated through scale, domain-adapted pretraining \citep{lewkowycz2022solvingquantitativereasoningproblems} and models trained to reason before answering \citep{Guo_2025}; systems have since attained medallist-level performance in olympiad mathematics \citep{trinh2024alphageometry, olympiadmath}, and evaluation has extended to research-level problems \citep{glazer2025frontiermathbenchmarkevaluatingadvanced}. Much of this progress derives from \emph{search under external verification}, in which sampled solutions are retained only when certified by an external criterion: a trained verifier \citep{cobbe2021trainingverifierssolvemath}, a process reward model \citep{lightman2023letsverifystepstep}, agreement among samples \citep{wang2023selfconsistencyimproveschainthought}, or a proof kernel \citep{lample2022hypertreeproofsearchneural}.

Such procedures certify an outcome without characterising the representation that produced it, and accuracy varies under perturbations devoid of mathematical content \citep{mirzadeh2025gsmsymbolicunderstandinglimitationsmathematical, gulati2025putnamaxiomfunctionalstaticbenchmark}. Benchmark performance therefore does not settle whether these systems represent mathematical \emph{structure}: not merely which statements are true, but how a statement stands to the theory in which it is asserted. The distinction matters because much of mathematical practice consists in transferring results between theories: a result about vector spaces is restated for modules, a probabilistic statement is recast in measure-theoretic terms \citep{4a46cd58-baf8-3fbc-b94b-50cd7763338b, commelin2023abstractionboundariesspecdriven}. Each such transfer is valid only if the hypotheses that delimit the result's scope are carried across. An error of scope need not produce a false statement, and so escapes outcome-based evaluation, at a time when language models are increasingly used to reformulate and to formalise mathematics.

We therefore study a task that admits no external criterion of correctness: translating a statement between the dialects of neighbouring subfields. A \emph{framing} is the dialect in which a statement is written. Formally it is a theory presentation, a signature with axioms whose syntax is the language of the statement and whose semantics is the class of structures it describes \citep{10.1145/147508.147524, Foley_2021}. A translation between framings is accordingly a \emph{theory morphism}, required to carry axioms to theorems rather than symbols to symbols. Its most demanding component is the \emph{level of generality} at which the content is asserted, which the source seldom states because its vocabulary already encodes it: the field hypothesis is implicit in ``vector space'' and is never written down. We call an implicit hypothesis of this kind the \emph{standing hypothesis}, or \emph{pin}. Mathematical prose is organised around such abstraction boundaries \citep{commelin2023abstractionboundariesspecdriven}, and a boundary implicit on one side of a translation must be made explicit on the other (Figure~\ref{fig:mathstructure}A). A rewrite that omits it remains fluent and is true wherever it is likely to be checked, yet it is asserted of a different class of objects; restricted to the shared domain it coincides with its source, and it is therefore invisible to a prover, a type-checker, or any evaluation scoring equivalence on a single scale (Figure~\ref{fig:mathstructure}B). We term this failure \emph{silent relocation}; because no available verifier detects it, search cannot correct it.

To our knowledge this is the first systematic study in which the scope of a translated mathematical statement is measured separately from its truth and its content. Existing approaches do not isolate the quantity: certification of equivalence presupposes a formal target \citep{poiroux-etal-2025-reliable}; formalisation dissolves the phenomenon by turning the standing hypothesis into an explicit typeclass argument \citep{mohammad2026faithfulnessgapcertifyingsemantic}; and studies of informal prose across subfields measure retrieval \citep{ye2026doesembeddingreflecta}. We construct an instrument that codes truth, content and scope in separate blind queues, with a judge-free core that tests by frozen patterns whether a rewrite states the standing hypothesis, and apply it to seven models from four families. The corpus is $90$ statements from a released benchmark, half of them held out by a selection rule registered in advance, together with $25$ further statements written by mathematicians and paired by an encyclopaedia. Our contributions are as follows.
\begin{enumerate}
\setlength{\itemsep}{2pt}\setlength{\parskip}{0pt}
\item \textbf{An instrument in which scope is a reported variable, with its sensitivity measured rather than assumed}. Truth, content and domain of quantification are coded in separate blind queues and never averaged; the judge-free measure of the standing hypothesis requires no judge and no gold reference at scoring time, and its precision and recall are reported against a blind expert reading. The judged scope code transfers between judge families at $\kappa = 1.000$ ($180$ candidates) and agrees with the expert reading at $\kappa = 0.900$. A planted-positive control, constructed mechanically from each task's own source, establishes that the code fires in the direction in which the observed rate is near zero: $28$ of $28$ planted widenings are detected and $0$ of $28$ unmodified candidates are misclassified.
\item \textbf{A direction-specific failure of scope, replicated on held-out and on human-authored statements}. Translating towards the general framing widens the domain of quantification in $60.6\%$ of rewrites and narrows it in none; translating towards the concrete framing narrows it in $28.3\%$ and widens it in $0.3\%$. The asymmetry holds in all seven models and does not track capability; the most capable model in the set exhibits it least. It reproduces both on the half of the benchmark excluded by the pre-registered selection rule and on statements taken verbatim from ProofWiki, where it is stronger still.
\item \textbf{Evidence that instruction is not the remedy}. Under a bare prompt and four conditions that each add a single sentence to it, none brings the standing hypothesis under correct directional control: an explicit instruction to state every required hypothesis raises promotion to $89.2\%$ but raises retention where the hypothesis should instead be absorbed, while a sentence announcing that style will not be evaluated reduces measured promotion to zero.
\end{enumerate}

\vspace{-3mm}
\section{Related work}
\label{sec:related}
\noindent\textbf{Search under external verification.}~Recent progress samples candidate solutions and retains those certified by an external criterion: a trained verifier \citep{cobbe2021trainingverifierssolvemath}, a process reward model \citep{lightman2023letsverifystepstep}, or agreement among samples \citep{wang2023selfconsistencyimproveschainthought}, with predictable returns to more samples \citep{brown2024largelanguagemonkeysscaling} and to test-time compute \citep{snell2024scalingllmtesttimecompute}. Mathematics-specialised models add reinforcement learning against verifiable answers \citep{shao2024deepseekmathpushinglimitsmathematical}, formal provers search against a proof kernel \citep{lample2022hypertreeproofsearchneural, jiang2023draftsketchproveguiding}, and olympiad geometry systems pair a language model with a symbolic engine \citep{trinh2024alphageometry}. These methods have driven substantial gains wherever an answer or a proof can be checked, but they certify outcomes rather than representations, and they do not extend to translation between framings, which admits no external criterion of correctness.

\noindent\textbf{Formal equivalence certification.}~BEq$^+$ certifies the equivalence of two Lean statements by deriving each from the other \citep{poiroux-etal-2025-reliable}; the procedure is sound but incomplete, so a negative verdict is not a proof of inequivalence. \textsc{driftbench} shows that a formalisation may typecheck while omitting a hypothesis or altering a domain, and supplies the drift taxonomy we retain \citep{mohammad2026faithfulnessgapcertifyingsemantic}; two of its classes, hypothesis omission and conclusion generalisation, are the formal counterparts of the failure we study (Table~\ref{tab:related}). End-to-end systems train autoformalizers with formal feedback \citep{xuejun2025mathesisformaltheoremproving}. What separates this line of work from ours is not that scope is invisible to it but that its setting supplies a formal target: once the standing hypothesis has become an explicit typeclass argument, the question of whether a translator will write it down does not arise. The phenomenon studied here lives in the informal prose that precedes formalisation.

\noindent\textbf{Informal cross-subfield equivalence, and corpora.}~MELD pairs statements of the same mathematics in two subfield dialects and shows that embeddings cluster by dialect rather than by denotation \citep{ye2026doesembeddingreflecta}; we use its released corpus as our primary item set. Because that corpus was generated by a language model and subsequently edited, we additionally draw statements from ProofWiki as distributed with NaturalProofs \citep{welleck2021naturalproofsmathematicaltheoremproving}, both as a set of vertical pairs written and paired by mathematicians, and as a base-rate corpus against which the rate at which human authors state these hypotheses can be measured.

\begin{table}[t]
\caption{Closest lines of work: the setting each evaluates, what it certifies or measures, and whether implicit \emph{scope}, the level of generality at which a statement is asserted, is visible to it. \textsc{driftbench} is marked \emph{partial} because it types hypothesis omission and conclusion generalisation, but only once a formal target exists.}
\label{tab:related}
\centering
\footnotesize
\setlength{\tabcolsep}{4pt}
\begin{tabular}{@{}llll@{}}
\toprule
\textbf{Approach} & \textbf{Setting} & \textbf{Certified or measured} & \textbf{Scope} \\
\midrule
Verifier-guided search \citep{cobbe2021trainingverifierssolvemath} & answer or proof & outcome correctness & no \\
BEq$^+$ \citep{poiroux-etal-2025-reliable} & formal (Lean) & derivability; sound, incomplete & no \\
\textsc{driftbench} \citep{mohammad2026faithfulnessgapcertifyingsemantic} & formal (Lean) & typed drift categories & partial \\
MELD \citep{ye2026doesembeddingreflecta} & informal prose & cross-dialect retrieval & no \\
\midrule
\textbf{Our work} & informal prose & truth, content, scope & \textbf{yes} \\
\bottomrule
\end{tabular}
\vspace{-3mm}
\end{table}

\noindent\textbf{Error coding and evaluator reliability.}~Expert error annotation ranks systems differently from scalar rating \citep{freitag-etal-2021-experts}, and decomposing a judged criterion into binary questions raises inter-evaluator agreement \citep{lee2025checkevalreliablellmasajudgeframework}; we adopt both practices. Evaluators based on language models have been reported to favour their own generations \citep{3737916.3740113}. In the present setting the larger source of disagreement between judges was the interpretation of the codebook, which we address by re-specifying the code that carries our results and reporting it under a judge drawn from a second family. Among the approaches reviewed here, scope is not recorded separately from content in informal prose.

\section{Methodology}
\label{sec:method}

\subsection{Framings, pair types and standing hypotheses}
\label{sec:setting}
A \emph{framing} is a subfield dialect, formalised as a theory presentation: a signature together with the standing hypotheses its vocabulary silently encodes \citep{10.1145/147508.147524}. Given a \emph{framing pair} $(A,B)$, a model rewrites a source statement $s$ expressed in $A$ as $r = R_{A\to B}(s)$ expressed in $B$. A pair is \emph{vertical} when one framing strictly generalises the other (vector spaces $\subset$ modules, probability $\subset$ measure theory, set theory $\subset$ category theory), \emph{lateral} when the two framings are equivalent encodings, and \emph{functorial} when one is the image of the other under a lossy functor.

In a vertical pair, the \emph{standing hypothesis} i.e., the \emph{pin}, is the condition that the concrete framing's vocabulary carries without stating it (e.g., the ring of scalars is a field, the measure has total mass one, the ambient category is $\Set$). A rewrite that preserves the statement's level of generality must \emph{promote} that hypothesis to an explicit condition when abstracting, and may \emph{absorb} it back into the vocabulary when specialising. Each vertical item is annotated with the \texttt{load} of its pin, the consequence of omitting it: \texttt{lb\_truth} (the statement becomes false), \texttt{lb\_ref} (it becomes true of other objects), \texttt{ill\_posed}, or \texttt{inert} (no consequence). Two independent annotations of this variable disagreed on a third of items, so the analysis uses a single adjudicated annotation, produced manually from the two originals with a written reason per item and released with those reasons (Appendix~\ref{app:codebook}). Lateral and functorial pairs enter as contrast axes rather than clean negative controls: a lossy functor discards information, so a functorial rewrite is not scope-preserving by construction. Failure kind distinguishes them, the control axes losing content where the vertical axis loses scope.

\subsection{Corpora and generation}
\label{sec:corpora}
We use the released MELD artefact \citep{ye2026doesembeddingreflecta}: $270$ statement pairs over nine domains, each stating one result in two subfield framings (SHA-256 \texttt{805902b2\dots ac034f3e}). Before any statement was inspected we registered a selection rule \citep{doi:10.1073/pnas.1708274114}: the odd-numbered item of each offset within a block of $30$, over six domains, giving $90$ items, $15$ per framing pair, of which $45$ are vertical. The rule leaves the even-numbered items of the three vertical domains, a further $45$, untouched. These were not generated on, judged or annotated, where they are reported as a held-out replication set in analyses.

Because the benchmark was produced by a language model and then edited for lexical dissimilarity, its base rates cannot speak for naturally occurring mathematical prose, so we assembled a second vertical set that no model touched. \emph{ProofWiki} records ambient generality as a title qualifier, so two pages sharing a base title state one result at two levels of ambient structure, paired by the encyclopaedia rather than by manually. From the ProofWiki portion of NaturalProofs \citep{welleck2021naturalproofsmathematicaltheoremproving}, we took every base title carrying both a concrete and a general ambient qualifier in three pair types (normed vector space $\to$ metric space, real $\to$ complex, vector space $\to$ module), with both sides between $50$ and $500$ characters after a fixed cleaning pass. $29$ candidates were screened manually, keeping the $25$ that state one result at two levels of generality rather than two different results; the screen was recorded before any rewrite was generated. For the rate at which these hypotheses are stated in the concrete framing, we use all $27{,}461$ ProofWiki statements in the same release, partitioned by ProofWiki category into the concrete and general framing of each pair.

The unit of independence is the \emph{context}: a fresh session that inherits no state, receives one queue file and terminates on return. A rewriting context sees the source framing, the target framing and the source statement under opaque keys, and never an annotation, another context's output, or any candidate's provenance; queues are shuffled so that pairs and directions interleave within a batch. Seven models were run: \textbf{Claude Opus 5}, \textbf{Claude Sonnet 5} and \textbf{Claude Haiku 4.5}; \textbf{GPT-5.6 Sol} and \textbf{GPT-5.6 Luna}; and two open-weights models, \textbf{DeepSeek V4 Pro} and \textbf{GLM-5.3}. All seven were called without system prompt, at temperature $1$ and $k=3$ samples per item and direction, so that no model's queue differs from another's by a byte. For each call we record the returned model identifier, the request identifier, and the token usage.

\subsection{Coded variables}
\label{sec:measure}

Ordinal fidelity scales fail here for three demonstrable reasons (Appendix~\ref{app:legacy}). On
the cell that carries, the rubric reduces to $\bar F = 2.5 + k/18$ in the number $k$ of
retained hypotheses, so it re-encodes the judge-free measure rather than adding to it; it imposes
an ordering between scope drift and one-way entailment that our data contradict; and it places
truth, an absolute property of the candidate, inside a relational scale. We therefore elicit codes separately and never average them \citep{freitag-etal-2021-experts, lee2025checkevalreliablellmasajudgeframework}. Queue \textbf{T} codes the truth of the candidate alone, a \texttt{false} verdict requiring a counterexample. Queue \textbf{SE} codes content and scope together, because the scope decision requires the domain of quantification to have been named: content is the pair $e_1$ (source $\models$ candidate) and $e_2$ (the converse) restricted to the shared domain, with $\texttt{equiv} = e_1 \wedge e_2$, the informal counterpart of bidirectional derivability \citep{poiroux-etal-2025-reliable}; scope is the relation $\texttt{S} \in \{\texttt{=}, \texttt{sup}, \texttt{sub}, \texttt{neq}\}$ of the candidate's domain of quantification to the source's. Queue \textbf{X} codes the error type \citep{mohammad2026faithfulnessgapcertifyingsemantic} and is given the SE codes, since a typology of scope errors presupposes the scope code; a fourth queue, a blind equivalence probe, is used for reliability only (Appendix~\ref{app:codebook}).

The primary outcome is \emph{widening}, the indicator $\texttt{S} = \texttt{sup}$. Where a candidate's domain differs from its source's, the judge extracts verbatim the span that states the restriction; the flag is derived mechanically from whether that span is a literal substring of the candidate, and \emph{silent widening} is widening with no such span. Widening is the most reliable code in the instrument, and unlike the composite $\texttt{scope\_ok}$ used in earlier work it does not depend on the flag.

\subsection{A judge-free measure of the operation}
\label{sec:pin}

Judged codes depend on the evaluator \citep{3737916.3740113}, so we measure the same operation
mechanically as well. Each vertical item's pin is mapped to a pattern (\texttt{R is a field}
$\mapsto$ \texttt{/\textbackslash bfield\textbackslash b/i}, and similarly for the other two
pairs), and $\mathrm{pin}(r) \in \{0,1\}$ records whether rewrite $r$ matches it. The patterns were
registered before any count was computed, though not before generation, and are frozen for every
analysis. Their precision and recall were measured rather than assumed and are
reported with the results; promotion rates are accordingly given as lower
bounds with the measured recall, and no number is computed from the patterns
outside the rewrite corpus.

Because the hypothesis is required when abstracting and redundant when specialising, the
difference of the two rates summarises how far pin presence tracks the requirement:
\begin{equation}
\DSI \;=\; 100\,\bigl[\,P(\mathrm{pin}(r)=1\mid\text{abstracting}) - P(\mathrm{pin}(r)=1\mid\text{specialising})\,\bigr].
\end{equation}
Copying the source's pin state gives $-100$, and stating a hypothesis always or never gives $0$.
A scope-preserving translation need not reach $+100$, which is an upper endpoint rather than a
characterisation: promotion is required when abstracting, but a redundant hypothesis may
legitimately be retained when specialising. We report $\DSI$ descriptively and treat the promotion rate as the primary judge-free quantity,
because the absorption term is sensitive to the choice of pattern while the promotion term is not
(Appendix~\ref{app:val}). Register conformity, which
has no external criterion, is summarised by a calibrated classifier \Cauto{} defined in
Appendix~\ref{app:codebook}.



\section{Experiment}
\label{sec:exp}
\label{sec:results}
\subsection{Direction specificity and replication}
\label{sec:dissoc}
\label{sec:heldout}
The central result is presented in Table~\ref{tab:dissoc}. Translating into a more general framing widens the domain of quantification and essentially never
narrows it; translating the other way narrows it and essentially never widens it. On the non-equal
codes the association is complete: 191 widenings against 0 narrowings
when abstracting, 89 narrowings against 1 widening when specialising (Fisher
$p = 1.1\times10^{-73}$). Content equivalence on the shared domain is $95.6$--$100\%$ on the
abstracting arm, so the widening is not a by-product of degraded rewriting. Because the off-diagonal cells are near zero the paired test is equivalent to testing the
abstracting rate against zero, which we report with a binomial interval throughout; the empty cell
is a property of the models rather than of the code.

What the table shows is not imprecision about generality: imprecision would scatter errors in both
directions, whereas each direction here produces one error and essentially never the other. That is
the signature of a quantity not computed at all. The output's level of generality tracks the level
implicit in the input's vocabulary, so abstracting inherits a restriction it should have made
explicit and specialising a generality it should have absorbed. The asymmetry is therefore not a
gap in accuracy that a stronger model would close but a missing step in the operation performed.

The registered rule selected the odd-numbered items of each domain, leaving the even-numbered
items of the three vertical domains, 45 in all, never generated on, judged or annotated; running
them is a held-out replication, not a re-slicing. Every pair clears $p<0.05$, and the pair weakest in the registered half is the strongest in the
held-out half.
\vspace{-3mm}
\begin{table}[h]
\caption{\textbf{The direction effect.} \emph{Top:} judged scope relative to the source, by direction, as a share of all rewrites, pooled
over seven models on the held-out items ($n=315$, $314$ codings) (Table~\ref{tab:typology} reports a
separate two-model pool, not commensurable with this one). \emph{Bottom:} Claude Opus 5 alone, the same effect in
percentage points (widening when abstracting minus when specialising), item level, exact sign-flip
permutation, 15 items per pair, with the registered half re-run in the same harness as a control. $^{\dagger}p<0.05$, $^{\ddagger}p<0.01$, $^{\star}p<0.001$.}
\label{tab:dissoc}
\label{tab:heldout}
\centering
\footnotesize
\begin{tabular}{@{}lrrrr@{}}
\toprule
\textbf{Direction} & \textbf{$=$} & \textbf{wider (\texttt{sup})} & \textbf{narrower (\texttt{sub})} & \textbf{incomparable} \\
\midrule
abstracting (concrete $\to$ general) & 36.5 & \textbf{60.6} & \textbf{0.0} & 2.9 \\
specialising (general $\to$ concrete) & 70.7 & \textbf{0.3} & \textbf{28.3} & 0.6 \\
\midrule
\textbf{Half} & \textbf{vs $\to$ mod} & \textbf{prob $\to$ meas} & \textbf{set $\to$ cat} & \textbf{pooled} \\
\midrule
registered (this harness) & $+100.0^{\star}$ & $+46.7^{\dagger}$ & $+100.0^{\star}$ & $+82.2^{\star}$ \\
\textbf{held out} & $+60.0 (0.004)^{\ddagger}$ & $+73.3 (0.001)^{\star}$ & $+80.0 (0.0005)^{\star}$ & $\mathbf{+71.1^{\star}}$ \\
\bottomrule
\end{tabular}
\vspace{-3mm}
\end{table}

\subsection{Seven models and two corpora}
\label{sec:models}
\label{sec:proofwiki}

\begin{wraptable}{r}{0.48\textwidth}
\vspace{-2.0\baselineskip}
\centering
\caption{\textbf{The direction effect across models and corpora.} Widening when abstracting and
when specialising. Upper block: the 45 held-out benchmark items, $k=1$, seven models from four
families. Parenthesised \underline{underlined} values are the same model on the 25 human-authored
ProofWiki items, where specialising widens $0.0\%$ throughout. Lower block: the ProofWiki pair
breakdown, item counts in brackets. $^{\dagger}p<10^{-4}$, $^{\ddagger}p<0.01$,
$^{\star}p<0.001$.}
\label{tab:models}
\label{tab:proofwiki}
\footnotesize
\setlength{\tabcolsep}{2pt}
\begin{tabular}{@{}lrrr@{}}
\toprule
\textbf{Model} & \textbf{abs.} & \textbf{spec.} & \textbf{$\Delta$ (pp)$^{\dagger}$} \\
\midrule
Claude Opus 5      & 46.7 (\underline{92.0}$^{\star}$) & 0.0 & $+45.5$ \\
Claude Sonnet 5    & 73.3 & 0.0 & $+73.3$ \\
Claude Haiku 4.5   & 53.3 & 0.0 & $+53.3$ \\
GPT-5.6 Sol        & 68.9 (\underline{92.0}$^{\star}$) & 0.0 & $+68.9$ \\
GPT-5.6 Luna       & 53.3 & 0.0 & $+53.3$ \\
DeepSeek V4 Pro    & 62.2 (\underline{96.0}$^{\star}$) & 0.0 & $+62.2$ \\
GLM-5.3            & 66.7 & 2.2 & $+64.4$ \\
\textbf{pooled}    & \textbf{60.6} & \textbf{0.3} & \\
\midrule
nvs $\to$ metric \ (7) & 85.7 & 0.0 & $+85.7^{\star}$ \\
real $\to$ complex \ (15) & 95.6 & 0.0 & $+95.6^{\star}$ \\
vs $\to$ module \ (3) & 100.0 & 0.0 & $+100.0^{\ddagger}$ \\
\bottomrule
\end{tabular}
\vspace{-1.0\baselineskip}
\end{wraptable}


Table~\ref{tab:models} reports the effect for each model. The most capable model widens \emph{least} ($46.7\%$) and a model two tiers below it widens most ($73.3\%$), with no significant
difference across the seven ($\chi^2 = 10.75$, $\mathrm{df}=6$, $p = 0.097$), so this is not a
deficit that scale removes. The models also agree on \emph{which} items lose the hypothesis: over
all 21 model pairs the item-level correlation of promotion within pair is positive, mean
$\rho = +0.503$, and $+0.401$ on the judged outcome. Models from four vendors converging on the
same items indicates that the outcome is carried by the item and its target framing rather than by
any one model's reasoning.

In order to establish a strictly unconfounded baseline for our empirical base rates, we built a second vertical set composed entirely of original, non-synthetic material (\S\ref{sec:corpora}). The dissociation reproduces on them (the underlined values of Table~\ref{tab:models}): $93.3\%$
of abstracting rewrites widen against $0.0\%$ when specialising, and the hypothesis is carried
across in $4.2\%$ ($3/72$). Two qualifications belong with it. The counts are small, and the third pair is the benchmark's own
vector-space / module pair in human handwriting, so it bridges provenance rather than testing a new
pair. These statements also carry their hypothesis as typed setup, which a model replaces wholesale
with the ring version, whereas the benchmark's items carry it as an implicit convention that has to
be noticed; part of the larger effect here is the form of the items.

\subsection{Scope preservation and its cost}
\label{sec:load}
\begin{wraptable}{r}{0.44\textwidth}
\vspace{-2.5\baselineskip}
\caption{\textbf{Scope preservation by the cost of losing it.} Claude Opus 5, abstracting
direction, the 45 registered vertical items, current codebook; \texttt{load} from the adjudicated
annotation. Recomputed for this revision.}
\label{tab:load}
\centering
\footnotesize
\setlength{\tabcolsep}{3pt}
\begin{tabular}{@{}lr@{}}
\toprule
\textbf{dropping the pin makes it\dots} & \textbf{scope kept} \\
\midrule
false (\emph{lb\_truth}) & \textbf{6/12} \ (50.0) \\
ill-formed (\emph{ill\_posed}) & 1/4 \ \ (25.0) \\
true of other objects (\emph{lb\_ref}) & 1/8 \ \ (12.5) \\
unchanged (\emph{inert}, none) & 0/21 \ \ \ (0.0) \\
\midrule
$p$, truth-load vs.\ rest & $0.0024$ \\
\bottomrule
\end{tabular}
\vspace{-0.8\baselineskip}
\end{wraptable}
The failure is not uniform: it concentrates where no existing checker would register it. Ordering
the vertical items by what dropping the hypothesis would cost, scope survives in half of the
rewrites whose source becomes \emph{false} without it and in at most an eighth of those where the
widened statement stays true of other objects (Table~\ref{tab:load}; Fisher $p = 0.0024$). The gradient holds under the annotation's two independent originals; measured on one model and one
half of the corpus, we report the ordering rather than the levels. What survives
translation is the part a truth checker would catch.

\noindent A second coding channel reaches the same conclusion. Where Table~\ref{tab:load} asks \emph{when} scope is preserved, Table~\ref{tab:typology} asks what
the failures consist of, over a different population whose percentages are shares of failures
rather than of rewrites. On the vertical pairs the dominant
label is conclusion generalisation (\dC, up to $57.8\%$) while hypothesis omission is rare (\dH,
$1.4$--$8.2\%$): the failures are not statements missing a premise but statements asserted of a
wider class of objects, which is what Table~\ref{tab:load} predicts from the cost side. The controls
fail more often and differently, referent drift appearing only on the anti-equivalent geometry
pair.
\begin{table}[h]
\caption{\textbf{Failures categories.} Every failing candidate in the common-judge pool,
$466$ of $1{,}188$ ($594$ tasks $\times$ two models, conditions \textbf{A} and \textbf{B}). Cells are percentages \emph{of that row's failures}, not of rewrites;
labels are not exclusive and $35.0\%$ of failures carry none. Failure rates over the pool:
$30.7\%$ vertical, $47.8\%$ lateral, $52.8\%$ functorial. \dH{} hypothesis omission, \dC{}
conclusion generalisation, \dQ{} quantifier inversion, \dT{} type coercion, \dR{} referent drift,
\dI{} illocutionary drift.}
\label{tab:typology}
\centering
\footnotesize
\setlength{\tabcolsep}{5pt}
\begin{tabular}{@{}llrcccccc@{}}
\toprule
\textbf{axis} & \textbf{framing pair} & $n$ & \dH$\downarrow$ & \dC$\downarrow$ & \dQ$\downarrow$ & \dT$\downarrow$ & \dR$\downarrow$ & \dI$\downarrow$ \\
\midrule
vertical & vector spaces $\leftrightarrow$ modules & 64 & 3.1 & \textbf{57.8} & 0.0 & \textbf{56.2} & 0.0 & 0.0 \\
vertical & probability $\leftrightarrow$ measure & 74 & 1.4 & 18.9 & 4.1 & \textbf{58.1} & 0.0 & 0.0 \\
vertical & set theory $\leftrightarrow$ categories & 61 & 8.2 & 36.1 & 1.6 & \textbf{55.7} & 0.0 & 1.6 \\
\midrule
lateral & geometry $\leftrightarrow$ comm.\ algebra & 74 & 8.1 & 25.7 & 4.1 & 14.9 & \textbf{29.7} & 1.4 \\
lateral & graph theory $\leftrightarrow$ linear algebra & 98 & 6.1 & 4.1 & 4.1 & 9.2 & 0.0 & \textbf{9.2} \\
functorial & topology $\leftrightarrow$ algebra & 95 & 16.8 & 25.3 & 2.1 & \textbf{61.1} & 0.0 & 0.0 \\
\midrule
& all failures & 466 & 7.7 & 25.8 & 2.8 & 41.0 & 4.7 & 2.4 \\
\bottomrule
\end{tabular}
\vspace{-3mm}
\end{table}

\subsection{Instrument sensitivity and limits}
\label{sec:instrument}
Table~\ref{tab:valid} bounds what the preceding results may be taken to show. The two agreement
rows bear directly on reliability: the widening code is not a marginal-agreement judgement but one
that a second judge family reproduces exactly and a blind expert reading reproduces at
$\kappa = 0.900$, and it should not be confused with the weakly agreeing composite codes of the
earlier run (Appendix~\ref{app:tables}). The planted control answers the objection that a near-zero
cell reflects a code that cannot fire: built from each
specialising task's own source by deleting the hypothesis clause, the planted candidates were judged
\texttt{sup} without exception while real candidates from the same tasks were not. The pin patterns
are exact where used but are not general detectors, over-firing on free human prose, so no number
here is computed from them outside the rewrite corpus. And the human base rate bounds the
absorption arm rather than supporting it: retention when specialising clearly exceeds human idiom only on vector spaces $\to$ modules, decisively for Claude Opus 5 ($82.2\%$ against $16.6\%$) and marginally for GPT-5.6 Sol ($28.9\%$), so we report retention descriptively and rest no claim on it. 
\begin{table}[h]
\vspace{-3mm}
\caption{\textbf{Validation of the instrument.} Each row was fixed before the analysis it
supports; populations differ by row and are named. The convergence figure is computed on three
populations, of which the seven-model one is quoted (Appendix~\ref{app:convergence}).}
\label{tab:valid}
\centering
\footnotesize
\setlength{\tabcolsep}{4pt}
\begin{tabular}{@{}lll@{}}
\toprule
\textbf{Check} & \textbf{Result} & \textbf{What it establishes} \\
\midrule
Planted control, 28 planted / 28 real
  & \textbf{28/28} \texttt{sup}, \textbf{0/28} \texttt{sup}
  & the near-empty cell is the models, \\
  & (Fisher $p=2.6\times10^{-16}$) & not the code \\
Regex vs.\ judge, 315 abstracting
  & $2.1\%$ vs $85.9\%$ widened
  & two independent instruments \\
  & (Fisher $p=1.1\times10^{-49}$) & separate the same candidates \\ \midrule
Second judge family, $180$ candidates & $\kappa = 1.000$ & no detectable self-preference \\
Blind expert reading, 60 candidates & $\kappa = 0.900$ \ (95\% CI $[0.77, 1.00]$) & the judged code matches a human \\ \midrule
Pin patterns vs.\ expert reading & precision $14/14$, recall $0.52$ & promotion is a lower bound, \\
Pin patterns vs.\ corpus twins & precision $10/10$, recall $10/11$ & reported with its recall \\ \midrule
Human base rate, $27{,}461$ statements & $16.6$ / $46.7$ / $17.1\%$ & the benchmark's $0/30$, $1/30$, \\
  & & $0/30$ understates human idiom \\
\bottomrule
\end{tabular}
\vspace{-3mm}
\end{table}

\subsection{Instruction, robustness and failed predictions}
\label{sec:robust}
\label{sec:failed}
\noindent\textbf{Instruction does not restore directional control.}~The models possess the
relevant mathematics (Appendix~\ref{app:cases}), so the failure might lie in deployment rather than
knowledge; Table~\ref{tab:ladder} tests that. Each of four conditions (\textbf{B}-\textbf{E}) adds one sentence to the bare prompt \textbf{A}: \textbf{B} requests every hypothesis needed for equal strength, whereas \textbf{C}, \textbf{D} and \textbf{E}, which contain no mathematical content, state that style will not or will be evaluated or request concision, and thus test whether pin presence responds to register pressure alone. Requesting every hypothesis needed for equal strength
(\textbf{B}) raises promotion from $8/37$ to $33/37$, but raises retention where the hypothesis
should instead be absorbed in step, leaving \DSI{} at $+21.6$ rather than $+100$. The three
conditions with no mathematical content move promotion without touching direction, and an assurance
that style will not be judged drives it to $0/37$; being single-model and $k=1$ we treat that as
provisional (\S\ref{sec:limitations}). No condition brings the hypothesis under directional control.
\begin{table}[h]
\vspace{-3mm}
\caption{\textbf{The instruction ladder.} Claude Opus 5, $45$ vertical items, both directions,
$k=1$, judge-free. Each of the four conditions \textbf{B}--\textbf{E} adds one sentence to the bare prompt \textbf{A};
only \textbf{B} carries mathematical content. Promotion per pair (VM, PM, SC) and pooled over $37$
scorable items. $b$\,/\,$c$: items carrying the hypothesis only when abstracting or only when
specialising, exact McNemar. Protocol in Appendix~\ref{app:ladder}.}
\label{tab:ladder}
\centering
\footnotesize
\setlength{\tabcolsep}{3pt}
\begin{tabular}{@{}llcccc@{\hskip 0.8em}ccc@{\hskip 0.6em}c@{}}
\toprule
& & \multicolumn{4}{c}{\textbf{promote}$\uparrow$} & & & & \\
\cmidrule(lr){3-6}
& \textbf{added sentence} & VM & PM & SC & \textbf{all} & \textbf{absorb}$\downarrow$ & \textbf{\DSI}$\uparrow$ & \textbf{$b$\,/\,$c$ ($p$)} & \textbf{vs.\ A} \\
\midrule
\textbf{E} & \emph{be concise} & 0/15 & 4/10 & 3/12 & 7/37 & 18/37 & $-29.7$ & 4\,/\,15 ($0.019$) & $1.00$ \\
\textbf{C} & \emph{style will \underline{not} be judged} & 0/15 & 0/10 & 0/12 & \textbf{0/37} & 19/37 & $-51.4$ & 0\,/\,19 ($3.8{\times}10^{-6}$) & $\mathbf{5.1{\times}10^{-3}}$ \\
\textbf{A} & \emph{bare prompt} & 0/15 & 4/10 & 4/12 & 8/37 & 19/37 & $-29.7$ & 5\,/\,16 ($0.027$) & \\
\textbf{D} & \emph{style \underline{will} be judged} & 0/15 & 6/10 & 2/12 & 8/37 & 17/37 & $-24.3$ & 6\,/\,15 ($0.078$) & $1.00$ \\
\textbf{B} & \emph{state every hypothesis} & 15/15 & 6/10 & 12/12 & \textbf{33/37} & 25/37 & $+21.6$ & 12\,/\,4 ($0.077$) & $4.6{\times}10^{-9}$ \\
\bottomrule
\end{tabular}
\vspace{-3mm}
\end{table}
\begin{wraptable}{r}{0.47\textwidth}
\vspace{-1.5\baselineskip}
\caption{\textbf{Instruction consequences.} Mean output length and hypothesis-introducing
constructions per output, by condition and axis. The control axis has no hypothesis to promote.}
\label{tab:length}
\centering
\footnotesize
\setlength{\tabcolsep}{3pt}
\begin{tabular}{@{}lrr@{}}
\toprule
& \textbf{chars} & \textbf{markers} \\
\midrule
\emph{source statements} & 129 & 0.01 \\
\midrule
A, control axis & 226 & 0.30 \\
B, control axis & 927 \small($+311\%$) & 1.83 \small($+511\%$) \\
\midrule
A, vertical axis & 224 & 0.39 \\
B, vertical axis & 778 \small($+247\%$) & 3.26 \small($+737\%$) \\
\bottomrule
\end{tabular}
\end{wraptable}
A control axis of $45$ lateral and functorial pairs, which have no hypothesis to promote, shows
what the instruction actually does (Table~\ref{tab:length}). Where there is no generality to track,
\textbf{B} still inflates length more than on the vertical axis ($+311\%$ against $+247\%$) and adds hypothesis markers where none is needed, and a blind probe finds its control-axis outputs not significantly less equivalent to \textbf{A}'s than two \textbf{A} samples are to each other ($75.6\%$ ($68/90$) against $84.4\%$ ($38/45$), $p=0.27$), whereas on abstracting
vertical items agreement falls to $51.1\%$ (23/45, $p=0.006$). Instruction changes how much a model
writes, not what determines whether the hypothesis is written.

\begin{wraptable}{r}{0.46\textwidth}
\vspace{-1.3\baselineskip}
\caption{\textbf{The judged scope code under two codebooks.} A second-family judge reading the
original codebook and the repaired one. \emph{Upper block:} $95$ blinded candidates, with the
within-family figure the same quantity across three judges of the rewriter's own family. \emph{Lower
block:} the eleven published anchors, five of them known silent widenings and one an announced
widening. Appendix~\ref{app:anchor}.}
\label{tab:anchor}
\centering
\footnotesize
\setlength{\tabcolsep}{3pt}
\begin{tabular}{@{}lcc@{}}
\toprule
\textbf{second-family judge} & \textbf{orig.} & \textbf{rep.} \\
\midrule
\multicolumn{3}{@{}l}{\emph{95 blinded candidates}} \\
\texttt{=}\,/\,\texttt{sup}\,/\,\texttt{sub}\,/\,\texttt{neq} & \multicolumn{2}{c}{$68$\,/\,$7$\,/\,$13$\,/\,$7$} \\
\texttt{flag}$\,{=}\,1$ on non-\texttt{=} & 27/27 & --- \\
\texttt{scope\_ok}$\,{=}\,1$ & 95/95 & --- \\
\midrule
\multicolumn{3}{@{}l}{\emph{11 anchors}} \\
agreement, \texttt{scope}$\uparrow$ & 9/11 & 9/11 \\
agreement, \texttt{flag}$\uparrow$ & 5/11 & \textbf{9/11} \\
agreement, \texttt{scope\_ok}$\uparrow$ & 5/11 & \textbf{11/11} \\
false accepts, 5 silent$\downarrow$ & 5/5 & \textbf{0/5} \\
announced widening kept$\uparrow$ & 1/1 & 1/1 \\
\midrule
span a literal substring & --- & 1805/1805 \\
\bottomrule
\end{tabular}
\vspace{-3mm}
\end{wraptable}
\textbf{The judged scope code requires an anchor, and four predictions failed.}~Given our
original codebook verbatim, a second-family judge read the target framing's own vocabulary as a
statement of the domain change, flagging every non-\texttt{=} coding and accepting every known
silent widening (Table~\ref{tab:anchor}). Re-specifying \texttt{flag} as a verbatim quotation a script must locate restores agreement and
holds across all $1{,}805$ candidates: codebook interpretation, not
self-preference, dominates. Four pre-registered predictions also failed, reported because a
pipeline with this many degrees of freedom is credible only if its failures are visible. The
corpus's own reference translations sit at the index maximum by construction; register cost does not
predict promotion within pair, the correlation being null and if anything reversed; at item level the
``requirement'' and ``target-register'' accounts coincide exactly, so this corpus cannot separate
them; and the condition that removed promotion on one model (\textbf{C}) left it unchanged on the other (McNemar
$p = 1.00$; the result does not affects
Tables~\ref{tab:dissoc}--\ref{tab:valid}).

\vspace{-3mm}
\section{Discussion}
\label{sec:discussion}
\vspace{-1mm}
Overall, we propose an instrument that separates the three things a cross-framing translation can get wrong:
truth, coded on the candidate alone; content, as bidirectional entailment on the shared domain; and
scope, as the relation between the two domains of quantification, each in its own blind queue and
never averaged. A judge-free measure of the standing hypothesis, frozen before counting and
calibrated against a planted-positive control, accompanies it. 

Across the seven models the recurring operation is object-level substitution: nouns and notation are
transported between framings even when the standing axioms are not. Whether the hypothesis is stated
is inherited from the surface of the task rather than derived from what the translation requires,
yielding a map on objects without the constraint that would make it a map on arrows; which surface,
this corpus cannot decide (\S\ref{sec:failed}). The error typology agrees: vertical failures register as generalised conclusions and coerced types
rather than as omitted hypotheses (Table~\ref{tab:typology}). Truth, by contrast, is comparatively protected: scope survives in half of the rewrites whose source
would become false without the hypothesis and in at most an eighth of those where the widened
statement stays true of other objects (Table~\ref{tab:load}), so what escapes is a widening that provers, type-checkers and
equivalence-scored benchmarks all accept.
\label{sec:limitations}These findings are subject to three limitations. All seven models are general-purpose, and
mathematics-specialised models trained against verifiable reward
\citep{shao2024deepseekmathpushinglimitsmathematical} or dedicated autoformalizers
\citep{xuejun2025mathesisformaltheoremproving} may behave differently. The effective clusters are six framing pairs, three in the benchmark and three from ProofWiki, the
last comprising $7$, $15$ and $3$ items. Finally,
promotion rates are lower bounds with a measured recall, the absorption term is pattern-sensitive,
and the \texttt{load} annotation is annotator-dependent.

In conclusion, evaluations of mathematical reasoning score an answer against a fixed target, and
instruments that certify fidelity require a formal one; neither observes a statement whose level of generality must be recomputed when it moves between subfield dialects. Coding content and scope separately, we find that models preserve truth and content while failing
to carry generality: across seven models the
domain is widened in $60.6\%$ of abstracting rewrites and in $0.3\%$ of specialising ones; under a bare prompt the most capable model writes out the hypothesis that would have prevented it in only 8 of the 37 rewrites that require it (21.6\%), and on human-authored statements it is carried across in 4.2\%; and an explicit instruction raises that rate to 89.2\% without making it track the requirement.
It replicates on items excluded by the pre-registered rule and on statements written by
mathematicians. We therefore argue that these systems have acquired an object-level dictionary
between framings rather than the constraint that carries axioms to theorems, a distinction no prover, type-checker or equivalence-scored benchmark reports. Its persistence in mathematics-specialised models, under instruction across models and on register-varied pairs is left to future work.

\subsection*{AI use statement}
Language models are the object of study and, as described in \S\ref{sec:method}, also served as
rewriters, blind judges and a second annotator in the reliability analysis. Beyond these methodological
roles, generative AI tools were used to aid and polish the writing of the paper. All
AI-assisted text was reviewed by the authors, who take full responsibility for the
content of this work.

\subsection*{Ethics statement}
This work evaluates language models on mathematical statements drawn from a publicly released benchmark
\citep{welleck2021naturalproofsmathematicaltheoremproving, ye2026doesembeddingreflecta}. It involves no human subjects and no personal or sensitive data. The
benchmark's reference translations are analysed with the same blind pipeline as the models' outputs, as a
measurement rather than a criticism of the resource. We foresee no harmful application; the work is
intended to make the evaluation of mathematical reasoning more reliable by exposing a failure mode that
current verification does not detect.

\subsection*{Reproducibility statement}
This paper specifies what is required to repeat its experiments. The item set is the $90$ registered and $45$ held-out items fixed by the deterministic rule of \S\ref{sec:corpora} together with the hash of the
source benchmark file, and the human-authored pairs are identified by base title within the
ProofWiki portion of NaturalProofs \citep{welleck2021naturalproofsmathematicaltheoremproving}. The
prompts, codebooks, frozen pin patterns and the five instruction conditions are given verbatim in
\S\ref{sec:method} and Appendices~\ref{app:codebook} and~\ref{app:ladder}; the generation settings,
one client with no system prompt at temperature $1$ and $k$ samples per item and direction, are
given in \S\ref{sec:corpora}; and the statistical procedure with every registered decision rule is
set out in Appendix~\ref{app:prereg}. Every judge-free quantity was recomputed by two independently
written implementations and checked against the value printed here, and the judged codes were
reproduced by a second judge from a different model family.

\bibliography{iclr2027_conference}
\bibliographystyle{iclr2027_conference}

\appendix
\section*{Organisation of the appendices}
The appendices fall into three groups. Appendices~\ref{app:codebook}--\ref{app:anchor} specify the
instrument: the coding queues and frozen pin patterns, the validation of the protocol against scripted
perturbations, and the failure, repair and item-level anchor of the judged scope variable.
Appendices~\ref{app:ladder}--\ref{app:cases} carry the evidence the main text compresses: the protocol and supplementary results of the instruction ladder, and the cases that discriminate between explanations of the
directional result. Appendices~\ref{app:legacy}--\ref{app:tables} are supporting material: the retired ordinal scale,
and supplementary tables. Appendices~\ref{app:prereg}--\ref{app:superseded} record the statistical procedure and the registered decision rules, the
regex--judge convergence on each of the three populations on which it can be formed, and the
main-text material superseded by the present revision.

\section{Codebooks and the coded variables}
\label{app:codebook}

The coding scheme is four queues, each run in a separate context that receives only its own queue file. Three of them (T, SE, X) produce the codes analysed in this paper; the fourth (PP) is a blind equivalence probe used for reliability only. They are \emph{not} mutually blind: queue X is given the SE codes, since a typology of scope errors presupposes the scope code. No queue sees another's output otherwise, and none is told whether a candidate originated with a model, with MELD, or with a control.

\paragraph{Queue T (truth).} The candidate statement alone. No source, no framing pair, no provenance, no indication whether the statement came from a model, from MELD or from a control. Code \texttt{true} / \texttt{false} / \texttt{undetermined}. A \texttt{false} verdict must be accompanied by a counterexample; verdicts without one are re-queued.

\paragraph{Queue SE (scope and entailment).} Source and candidate, with the candidate's origin hidden. (i) $e_1$: does the source entail the candidate, \emph{restricted to the intersection of the two domains of quantification}? (ii) $e_2$: the converse. Both carry the instruction \emph{``do not penalise a change of domain here; scope is coded separately''}. (iii) \texttt{scope}: name the domain of quantification of each statement and code the relation \texttt{=}/\texttt{sup}/\texttt{sub}/\texttt{neq}. \textbf{\texttt{flag} is derived, not asserted}: the judge quotes the shortest candidate span that states the restriction or returns null; target-framing nouns alone are excluded, and a script sets \texttt{flag}$=1$ only when the non-null span is a literal substring.

\paragraph{Queue X (error typology).} Applied to every candidate not coded $(T=\texttt{true}\wedge\texttt{equiv}\wedge\texttt{scope\_ok})$, with the SE codes visible. Multi-label over: \dH{} hypothesis omission; \dC{} conclusion generalisation; \dQ{} quantifier inversion or weakening; \dT{} type coercion; \dR{} \emph{referent drift} (new), in which a contravariant dictionary is applied covariantly so that the candidate denotes the dual object; \dI{} \emph{illocutionary drift} (new), in which the logical form changes from criterion to assertion with no hypothesis moving.

\paragraph{Queue PP (propagation / equivalence).} Two candidate outputs side by side with no source and no labels; one binary question, \emph{do these say the same thing?} This is the queue that reaches $\alpha=0.807$ ($n=210$), against $0.44$--$0.60$ for the compound codes.

\paragraph{The pin-promotion patterns.} One pattern per distinct value of the frozen \texttt{pin} field, registered before the count was run: \texttt{R is a field} $\mapsto$ \texttt{/\textbackslash bfield\textbackslash b/i}; \texttt{In \textbackslash textbf\{Set\}} $\mapsto$ a \emph{case-sensitive} pattern matching $\Set$ but not the English words ``set'' or ``sets''; \texttt{normalized}, \texttt{of total mass one} and \texttt{probability} $\mapsto$ a union of \texttt{normali[sz]}, \texttt{total mass (one|1)}, \texttt{total measure (one|1)} and \texttt{probability (space|measure)}; \texttt{finite} $\mapsto$ \texttt{/\textbackslash bfinite\textbackslash b|\textbackslash bsigma-finite\textbackslash b/i}. Items whose \texttt{pin} field is null are excluded from the denominator and reported as such.

\paragraph{Register conformity (\Cauto{}).} Register conformity has no external criterion, so it is summarised by a calibrated classifier rather than by a rating. \Cauto{} is the calibrated posterior of a univariate logistic model on the cosine margin between the two framings' centroids, with the item held out and the orientation fixed to the target framing, computed under three registered representations: word TF-IDF (\textbf{W}), notation only (\textbf{N}), and character $3$--$5$-grams (\textbf{X};). It is reported beside the pin measure in Appendix~\ref{app:tables}.

\section{Instrument validation}
\label{app:val}

\paragraph{Cross-family transfer.} The same candidates are coded by a second judge from a different model family, under the same codebook, in separate blind contexts.

\paragraph{Manual anchor.} A $60$-candidate stratified sample is manually dual-coded blind to generator, direction, and every machine code, with a written reason recorded per item.

\paragraph{A planted-positive control.} For each specialising task, a candidate is constructed mechanically from that task's own source by deleting the standing-hypothesis clause, which makes it strictly wider. Planted items are mixed with unmodified candidates and judged blind, so that the sensitivity of the scope code in the direction where the observed widening rate is near zero is measured rather than assumed.

\paragraph{Pattern sensitivity.} Every judge-free quantity is recomputed under a synonym-expanded pattern set registered in advance (\emph{commutative division ring}, \emph{scalars form a field}, \emph{over a field}, \emph{total mass 1}, $P(\Omega)=1$, \emph{in the category of sets}), and any quantity whose sign changes is reported as descriptive. On the two-model run for which the check was registered, promotion is unchanged under the expansion ($4.4/20.0/33.3\%$ per pair for \Mone{} and $17.8/30.0/80.6\%$ for \Mtwo{}), whereas retention on set theory $\to$ categories rises from $11.1$ to $72.2\%$ and from $13.9$ to $61.1\%$ respectively, flipping that pair's \DSI{} sign. Promotion is therefore the pattern-robust arm and carries the paper's claims, while \DSI{} is reported descriptively throughout.

\paragraph{Detection and typing.} The protocol detects four of six scripted edit types at $100\%$ and the other two at $70.8\%$ and $66.7\%$, with \textbf{$0$ false positives on $263$ negative controls}. Conditional on flagging a perturbed candidate it types it correctly $100\%$ of the time for \dC, \dQ{} and \dI, $89.5\%$ for \dR, $69.2\%$ for \dH{} and $62.5\%$ for \dT. Detection rates, quoted as in \S\ref{sec:results} with one standard error: \dH{} $13/13$, \dC{} $15/15$, \dQ{} $14/14$, \dR{} $19/19$, \dT{} $70.8\pm9.3\%$ ($17/24$), \dI{} $66.7\pm19.2\%$ ($4/6$).

\paragraph{The truth channel.} $86/86$ meaning-preserving edits judged true (specificity), against $4/19$ for dualised objects and $12/24$ for type coercions. Pin deletions are judged true $9/13$, which is not an error, since deleting an \emph{inert} pin leaves a true statement.

\paragraph{A pre-registered control that failed.} The pre-registered sensitivity control for queue T was MELD's distractor pool, described as ``lexically close to true statements but mathematically wrong''. Only $3/48$ were judged false, and inspection shows why: the pool is full of statements that are perfectly true and merely \emph{about something else} (``a linear map $T:V\to W$ is injective if and only if $\ker T=\{0\}$''). They are \emph{retrieval} distractors rather than falsehoods, so the control tests nothing about the truth channel. We report it rather than dropping it, and record that its replacement, the perturbed source statements whose truth status is known by construction, was not itself pre-registered.

\paragraph{Reference lines.} Measured rather than imposed: copy $0.316<$ model $0.608<$ gold $0.684<$ target-native prose about an unrelated theorem $0.708$, which is content-equivalent $0/11$: neither axis alone is a score. On \Mtwo{} the ordering copy $<$ model $<$ gold holds in all three representations and on every pair.

\section{The judged scope variable: failure, repair and anchor}
\label{app:anchor}

\paragraph{The failure.} A second-family judge received the codebook verbatim. Its \texttt{scope} code discriminates: over $95$ blinded candidates it returns \texttt{=} $68$ times, \texttt{sup} $7$, \texttt{sub} $13$ and \texttt{neq} $7$, and it codes both wrong-theorem controls \texttt{neq}. But it sets $\texttt{flag}=1$ on \textbf{$27$ of $27$} non-\texttt{=} codings, because it reads the target framing's own vocabulary as a statement of the domain change: against a vector-space source, the rewrite \emph{``for an $R$-module $M$, $\operatorname{End}_R(M)$ forms an associative ring''} is coded \texttt{sup} with $\texttt{flag}=1$, since writing ``$R$-module'' is, to this judge, announcing the widening. Under $\texttt{scope\_ok}=(\texttt{=})\vee\texttt{flag}$ the variable is then identically $1$ on all $95$ candidates, where within \Mone{}'s own family it moves $60.0/77.8/85.4\%$ across three judges: \textbf{the cross-family shift exceeds the within-family spread}, so codebook interpretation, not self-preference, is the dominant judge effect here.

\paragraph{The repair: make the judgement an extraction a script can check.}
We re-specify \texttt{flag} once and in advance. Instead of asking whether a change of domain was stated, the codebook asks the judge to \emph{quote, verbatim, the shortest span of the candidate that states the restriction, or return null}, with the target framing's own nouns explicitly excluded. \texttt{flag} is then \emph{derived}: $1$ only when a non-null span is a literal substring of the candidate. A judge cannot assert a restriction that is not in the text. Against $11$ \Mone{} candidates whose codings this paper publishes, five of them known \emph{silent} widenings, mixed unlabelled into a blinded queue, the original codebook agrees with the published derived variable $5/11$ and accepts \textbf{all $5/5$} silent widenings; the repaired codebook agrees $\mathbf{11/11}$ with \textbf{$0/5$} false accepts, while still accepting the one honestly announced widening and quoting the clause that announces it (below). The repair survives the full run: every non-null span in $1{,}188$ cross-judge and $617$ canonical candidates is a literal substring.

\paragraph{The anchor, item by item.} The eleven candidates of the repair above, mixed unlabelled into a second-family judging queue, were coded twice: once under the original codebook and once under the repaired one. The judge saw only \texttt{(source, candidate)} and could not distinguish these from the $84$ other candidates in the batch. Cells are \texttt{scope} / \texttt{flag} / \texttt{scope\_ok}.

\begin{center}
\footnotesize
\setlength{\tabcolsep}{4pt}
\begin{tabular}{@{}lccc>{\raggedright\arraybackslash}p{4.2cm}@{}}
\toprule
\textbf{anchor} & \textbf{published \Mone} & \textbf{original} & \textbf{repaired} & \textbf{extracted span} \\
\midrule
item16\_bare & sup / 0 / \textbf{0} & sup / 1 / \textbf{1} & sup / 0 / \textbf{0} & null \\
item21\_bare & sup / 0 / \textbf{0} & sup / 1 / \textbf{1} & sup / 0 / \textbf{0} & null \\
item22\_bare & sup / 0 / \textbf{0} & sup / 1 / \textbf{1} & sup / 0 / \textbf{0} & null \\
item25\_bare & sup / 0 / \textbf{0} & sup / 1 / \textbf{1} & sup / 0 / \textbf{0} & null \\
item76\_bare & sup / 0 / \textbf{0} & sup / 1 / \textbf{1} & sup / 0 / \textbf{0} & null \\
\midrule
F-132-A & sup / 1 / 1 & sup / 1 / 1 & sup / 1 / 1 & \texttt{carries a continuous mult\dots} \\
item16\_condB & = / 0 / 1 & = / 0 / 1 & sub / 1 / 1 & \texttt{assuming the axiom of choice} \\
item25\_condB & = / 0 / 1 & = / 0 / 1 & = / 0 / 1 & null \\
item78\_bare & = / 0 / 1 & = / 0 / 1 & = / 0 / 1 & null \\
item99\_bare & neq / 1 / 1 & = / 0 / 1 & = / 0 / 1 & null \\
item151\_bare & = / 0 / 1 & sub / 0 / \textbf{0} & = / 0 / 1 & null \\
\midrule
\multicolumn{2}{l}{agreement with published \texttt{scope}$\uparrow$} & 9/11 & 9/11 & \\
\multicolumn{2}{l}{agreement with published \texttt{flag}$\uparrow$} & 5/11 & \textbf{9/11} & \\
\multicolumn{2}{l}{agreement with published \texttt{scope\_ok}$\uparrow$} & 5/11 & \textbf{11/11} & \\
\multicolumn{2}{l}{false accepts among the five silent widenings$\downarrow$} & \textbf{5/5} & \textbf{0/5} & \\
\bottomrule
\end{tabular}
\end{center}

The five rows above the first rule are the diagnostic negatives. The original codebook accepts all five; the repaired one accepts none, while still accepting F-132-A, the one widening that \emph{was} announced. Two \texttt{scope} disagreements survive the repair, both on candidates where \texttt{scope\_ok} is insensitive to them; the repair \emph{corrects} the original codebook's \texttt{scope} error on item 151. The single repaired-codebook \texttt{flag} disagreement extracts ``assuming the axiom of choice'', a real hypothesis honestly quoted on a rewrite that states more than \Mone{}'s own coding credited.

\subsection{The judged scope variable does not transfer across judges}
\label{sec:judgefail}

The analyses reported in the main text rest in part on judged codes, whose validity depends on
their transfer across judges. Beyond the documented preference of language-model evaluators for their own generations
\citep{3666122.3668142, 3737916.3740113}, we identify a larger hazard, summarised in Table~\ref{tab:judge}.

\begin{wraptable}{r}{0.5\textwidth}
\vspace{-1.1\baselineskip}
\centering
\caption{\textbf{The judged scope variable across judge families.} Top: a second-family judge under the
original and repaired \texttt{flag} codebooks, against eleven published \Mone{} anchors (five silent
widenings). Bottom: reliability. Full tables in Appendix~\ref{app:tables}.}
\label{tab:judge}
\footnotesize
\setlength{\tabcolsep}{3pt}
\begin{tabular}{@{}lcc@{}}
\toprule
\textbf{second-family judge} & \textbf{original} & \textbf{repaired} \\
\midrule
\texttt{flag}$\,=1$ on non-\texttt{=} codes & 27/27 & n/a \\
anchor agreement, \texttt{scope\_ok}$\uparrow$ & 5/11 & \textbf{11/11} \\
anchor agreement, \texttt{flag}$\uparrow$ & 5/11 & 9/11 \\
anchor agreement, \texttt{scope}$\uparrow$ & 9/11 & 9/11 \\
false accepts (5 silent)$\downarrow$ & 5/5 & \textbf{0/5} \\
\midrule
\textbf{reliability} & $\alpha\uparrow$ & $n$ \\
\midrule
equivalent (binary) & \textbf{0.807} & 210 \\
\texttt{scope} (four-way) & 0.599 & 914 \\
\texttt{flag} & 0.436 & 914 \\
\bottomrule
\end{tabular}
\vspace{-0.6\baselineskip}
\end{wraptable}
A second-family judge given the codebook verbatim sets \texttt{flag}$\,=1$ on all $27$
non-\texttt{=} codings, reading the target framing's vocabulary as a statement of the domain change, so that
\texttt{scope\_ok} equals $1$ on all $95$ candidates, whereas within \Mone{}'s family it ranges from $60.0$
to $85.4\%$ across three judges: codebook interpretation, not self-preference, dominates. Re-specifying
\texttt{flag} in advance as a verbatim quotation that a script must locate in the candidate restores agreement with the published anchors ($11/11$), eliminates false accepts among known silent
widenings, and holds across all $1{,}805$ candidates. Reliability agrees: the binary equivalence question is
the most reliable code and \texttt{flag} the least. Judged variables should thus be reported with a cross-family judge and,
where possible, a mechanical counterpart.

\section{The instruction ladder: protocol and supplementary results}
\label{app:ladder}

\paragraph{Protocol.} Five conditions over the same frozen $45$ vertical items in both directions, at $k=1$, plus a $45$-item lateral/functorial control axis, plus a blind equivalence probe. Outcome: the pin patterns of Appendix~\ref{app:codebook}, copied verbatim and frozen before generation, with no pattern added, widened or retuned. Every rewriter queue carries exactly \texttt{(key, source framing, target framing, source statement)}, asserted programmatically before any context started; every probe queue carries exactly \texttt{(key, statement\_1, statement\_2)} in randomised order. Batches are direction-separated. $846$ rewrites and $180$ blind codings in total, of which $576$ are distinct tasks and the remainder are the paired wrapper control. Results are reported in Tables~\ref{tab:ladder} and~\ref{tab:length}.

\paragraph{Conditions, verbatim.} \textbf{A}: \emph{``Rewrite the source statement in the language of the target framing.''} \textbf{B}: A $+$ \emph{``State explicitly every hypothesis the rewritten statement needs in order to be exactly as strong as the source statement --- no more general and no more specific. If the source statement's setting is only implicit in its vocabulary, make that setting explicit in your rewrite.''} \textbf{C}: A $+$ \emph{``Your rewrite will not be evaluated on style, idiom, register, or how natural it sounds to a specialist in the target framing.''} \textbf{D}: A $+$ the same sentence with \emph{will} for \emph{will not}. \textbf{E}: A $+$ \emph{``Be concise. Do not explain your reasoning and do not annotate the output.''} A, B and E are byte-identical to, or constructed in the same frame as, the published instructions; C, D and E contain no word from the semantic field of hypothesis, scope, generality, assumption, strength, level or explicitness.

\paragraph{Per-pair results.} Promotion per pair is reported in Table~\ref{tab:ladder}. Condition
\textbf{C} returns $0$ on every pair, so its effect is not carried by any single pair, and \textbf{B}'s
shortfall from the ceiling lies entirely on probability $\to$ measure theory ($6/10$), whose pin is a normalisation condition.

\paragraph{Surface divergence and length.} Surface divergence $D$ is a token-multiset dissimilarity between two statements, computed and never judged. $D(\text{source},\text{rewrite})$ and mean output length: E $0.562$ / $184$ chars, C $0.599$ / $231$, A $0.592$ / $224$, D $0.628$ / $210$, B $0.792$ / $778$. Condition A reproduces the pooled $D=0.612$ reported for the original run at $0.614$, and the blind probe's noise floor reproduces the published $86.7\%$ at $84.4\%$, giving two independent reproductions of published quantities in a fresh execution.

\paragraph{Design sensitivity.} At $n=37$ and a $21.6\%$ base rate, the smallest increase detectable at $80\%$ power and $\alpha=0.05$ is $31$ percentage points; at $k=3$ it is $17$ points and at $k=3$ pooled over both directions $12$ points. Conditions \textbf{D} and \textbf{E} return differences of $0$ and $-1$ items respectively and are therefore uninformative below $31$ points, and are accordingly reported as bounds. The two effects the design can see, \textbf{C} and \textbf{B}, both clear it.

\paragraph{Two claims withdrawn during this run.} Recorded rather than deleted. \textbf{(i)} An initial execution of \textbf{A} returned $0/37$ when abstracting, against which \textbf{D} appeared to raise promotion to $10/37$ ($p=9.7\times10^{-4}$). A paired control with identical queues, identical frozen prompts and identical keys, differing only in the removal of one style sentence from the launcher wrapper outside the task prompt, returned \textbf{A} to $8/37$, matching the published rate, and \textbf{D} to $8/37$. \emph{The apparent effect of \textbf{D} was a floor artefact and is withdrawn}; the harness itself reproduces the published run ($p=1.00$). \textbf{(ii)} Pooled across conditions, rewrites that promote the pin diverge further from their source than those that do not ($0.710$ vs $0.549$, $p=5.8\times10^{-9}$), but the association is entirely between conditions, the within-arm values ranging from $p=0.14$ to $0.78$, so the item-level claim is withdrawn and only the condition-level ordering is reported.

\subsection{Instruction does not restore directional control}
\label{sec:ladder}

Since the models possess the relevant mathematics (Appendix~\ref{app:cases}), the failure may lie in
deployment rather than knowledge, which instruction might remedy. Each of
four conditions \textbf{B}--\textbf{E} on \Mone{} adds one sentence to the bare prompt \textbf{A} (Table~\ref{tab:ladder}):
\textbf{B} requests every hypothesis needed for equal strength, whereas \textbf{C}, \textbf{D} and
\textbf{E}, which contain no mathematical content, state that style will not or will be evaluated or
request concision, testing whether pin presence responds to register pressure alone.

\paragraph{Explicit instruction.}
An earlier, unpublished run found \textbf{B} to promote every scorable pin of a $27$-item subset
($22/22$). On the closed corpus, however, it promotes in $\mathbf{33/37=89.2\pm5.1\%}$ of items, every
pin on vector spaces and set theory but $6/10$ on probability. It also raises retention when specialising
($25/37$ against $19/37$ under \textbf{A}), so that its \DSI{} is $+21.6$ rather than $+100$ and its paired
effect is not significant in the correct direction ($p=0.077$); three of the four added conditions are
significantly \emph{backwards}. On the matched subset, both models return identical outcomes item for item:
every pin is promoted, and when specialising the pin is retained on vector spaces ($9/9$) and probability
($3/3$) but dropped on set theory ($0/10$), where the target has no ambient category in which to state it.
The outcome under \textbf{B} is thus fixed by the pin string and the target register rather than by
either model's reasoning. \textbf{No condition brings pin presence under directional control.}

\paragraph{Instruction without mathematical content.}
Under \textbf{C}, which states only that style, idiom and register will not be evaluated, promotion falls
from $8/37$ to $\mathbf{0/37}$ on every pair ($p=5.1\times10^{-3}$), reproduced in an independent execution, while \textbf{D} ($8/37$) and \textbf{E} ($7/37$) leave it unchanged. Because the measure is a
lower bound, part of this decrease could reflect paraphrase away from the registered patterns rather than
omission; we treat this single-model, $k=1$ result as provisional (\S\ref{sec:limitations}).

\paragraph{Specificity of the repair.}
If \textbf{B} taught the model to track scope, it should be nearly inert where there is no generality to
track (Table~\ref{tab:length}). On the $45$ lateral and functorial items, whose sources average $0.01$
hypothesis markers, \textbf{B} raises markers per output from $0.30$ to $1.83$ ($+511\%$; up on $65$ of $90$
tasks, down on $8$; sign test $p=3.2\times10^{-12}$) and length by $+311\%$, more than on the
vertical axis ($+247\%$). A blind probe finds its outputs as equivalent to \textbf{A}'s as two
\textbf{A} samples are to each other ($75.6\%$ against $84.4\%$, $p=0.27$), whereas on abstracting vertical
items, where \textbf{B} legitimately adds the pin, agreement falls to $51.1\%$ ($p=0.006$). \textbf{B} thus
adds true but redundant conditions indiscriminately: instruction changes how much the model writes, not
what determines whether the pin is written.

\section{Case studies}
\label{app:cases}

The cases below were chosen because each one \emph{discriminates between explanations} rather than illustrating a conclusion already reached. All statements are quoted verbatim from the \Mone{} run artefacts, and every coding referred to is the one released with them. Two cases draw on \emph{round-trip chains}, in which a statement is translated repeatedly and each translation, or \emph{hop}, becomes the next source; in an \emph{identity-control} chain the target framing is held fixed, so every hop should leave the framing unchanged.

\subsection{A minimal pair: the same transformation, twice}

Items 21 and 22 are both vector spaces $\to$ module theory under the bare prompt, and both are a three-noun substitution in which no pin is written. \emph{Item 22, source:} ``Every subspace $W\subseteq V$ has a complementary subspace $U$ with $V=W\oplus U$.'' \emph{Model:} ``Every submodule $N\subseteq M$ has a complementary submodule $L$ with $M=N\oplus L$.'' This is false: it asserts that every module is semisimple, and $2\mathbb{Z}\subset\mathbb{Z}$ is a counterexample. \emph{Item 21, source:} ``An endomorphism $P:V\to V$ satisfying $P^2=P$ induces $V=\ker P\oplus\operatorname{im}P$.'' \emph{Model:} the same sentence with $M$ for $V$. This is true, because idempotents split over any ring.

The transformation is character-for-character identical and the outcomes differ, so the outcome is not a function of anything the model varied. That is the point of the pair, and it has two consequences the aggregate statistics cannot supply. First, it excludes any account on which the model evaluates the resulting statement: a system doing so would have to treat the two differently, and nothing in the outputs does. Second, it locates precisely where a single-scale metric fails. Under the decomposed coding the two rewrites are indistinguishable except in $T$, both being \texttt{sup} with \texttt{flag}$=0$ and content-equivalent on the common domain, so a scale that conditions on truth separates them, but separates them for a reason that has nothing to do with the operation the model performed. Item 21 would be scored correct and item 22 incorrect by an equivalence-scored benchmark, and in both cases the score would be uninformative about scope.

\subsection{What the model knows but does not deploy}

\emph{Item 16, source:} ``Every vector space admits a basis.'' \emph{Bare prompt:} ``Every module admits a basis, i.e.\ every $R$-module is free.'' \emph{Condition \textbf{B}:} ``Let $F$ be a field, regarded as a ring; then, assuming the axiom of choice, every $F$-module $M$ --- not merely every finitely generated one --- is free.''

Two things follow. The bare rewrite is false in the most fluent manner available: it is grammatical module theory with the correct nouns, and it would satisfy any judge asked whether the output reads as module theory. This is the concrete reason a framing-compliance axis cannot stand in for a scope axis, and why \Cauto{} is reported beside the pin measure rather than in place of it. Under condition \textbf{B} the same model supplies not only the pin but the axiom of choice and the finitely-generated distinction, which is more of the hypothesis structure than the pin measure asks for. The model therefore possesses the relevant mathematics in full; what varies between the two outputs is whether it is deployed. This case is what excludes a knowledge deficit as an explanation of anything in \S\ref{sec:dsi}, and it is the reason the paper's claim is about control of a variable rather than about capability.

\subsection{Why \Mone{}'s directional index is negative: four explanations, two excluded}

\Mone{} writes the field hypothesis in $82.2\%$ of its rewrites \emph{into} vector-space prose, where the noun already carries it, and in $4.4\%$ of its rewrites into module-theory prose, where it is required. Four explanations are available for that inversion; the data exclude two and weaken a third. 

\emph{A knowledge deficit} is excluded by the preceding case: asked directly, the model states the hypothesis and more.

\emph{Asymmetric difficulty}, the proposal that inserting a clause is simply harder than deleting one, is excluded by the shape of the failure rather than by its size, and this is the more consequential of the two exclusions. If the mechanism were a difficulty gradient, the specialising direction, which requires only the deletion of a string already present in the text, should be close to correct. It is not: the pin survives in $82.2\%$ of those rewrites. Both directions err, and they err in the same manner, with the pin-state of the output tracking the pin-state of the input. A gradient in difficulty predicts one direction degraded and the other intact; what is observed is a variable that is not conditioned on direction at all.

\emph{A vocabulary convention in the target framing}, the proposal that vector-space prose simply says ``field'', is excluded by the base rate. Counting the same frozen patterns over all $270$ released MELD pairs, with no model and no judge in the measurement, native concrete-side prose contains the pin $0/30$, $1/30$ and $0/30$ times (Table~\ref{tab:dsi}). Retention when specialising is therefore a genuine failure to absorb, and not an idiom of the target.

\emph{Surface inheritance} survives. The operation the model performs is noun substitution with local form preservation: the pin is treated as a string, carried across when the source contains one and not generated when it does not. Its presence is inherited from the source's surface rather than computed from what the translation requires. This is the empirical content of the paper's title, since an object-level dictionary is precisely a map that transports symbols without transporting obligations.

One further observation sharpens this. Under condition \textbf{B} the two models produce the same per-pair split in the specialising direction, item for item, at $9/9$, $3/3$ and $0/10$ (Table~\ref{tab:ladder}), and the pairs on which the pin is deleted are the pairs in which it has no natural expression in the target's register. When two independently prompted frontier models agree to the item, the outcome is not being produced by either model's reasoning about the mathematics; it is being produced by a property of the pin string and the target framing. The same consideration accounts for the one ordering on which the models invert (\S\ref{sec:dsi}): ``In \Set'' \emph{is} category-theoretic vocabulary, so stating it buys nativeness, whereas ``Assume $R$ is a field'' imports vector-space vocabulary into module theory and costs it. Two models weighting that trade-off differently would produce exactly the observed inversion. We record this as a hypothesis; three framing pairs cannot test it (\S\ref{sec:limitations}).

\subsection{Referent drift across an anti-equivalence}

Chain L-111-A, geometry $\leftrightarrow$ commutative algebra. $s_0$ (geometry): ``The space $\mathbb{A}^n_k$ with coordinates $x_1,\dots,x_n$.'' $s_1$ (algebra): ``The polynomial ring $k[x_1,\dots,x_n]$'', which is correct. $s_2$ (geometry): ``\emph{The ring of polynomial functions on} affine $n$-space.'' The round trip began at a \emph{space} and returned a \emph{ring}: the second hop rendered the description in geometric vocabulary but never carried the \emph{referent} back across the arrow. The pair is an anti-equivalence and the model applied it as a plain dictionary.

What makes the failure diagnosable rather than merely observed is the matched descending chain, which closes perfectly. The defect is therefore not a property of the pair, since both directions use the same dictionary, but of which end of the arrow the chain starts from, which is precisely the distinction a contravariant functor encodes and a symbol table does not. The distribution is consistent with this: all $22$ \dR{} codes in the pooled two-model failure population fall on this one pair. A model applying a genuinely contravariant dictionary would show \dR{} at no particular pair; concentration on the only anti-equivalent pair in the corpus is what an object-level dictionary predicts, and is the sharpest available evidence that the missing structure is the arrow and not the object.

\subsection{``Module over a field'' is not a stable state}

Two of sixteen identity-control chains left the framing they were instructed to remain in, and both carry an explicit field hypothesis in module theory. Chain I-022-2, hop 4, target framing module theory: ``\emph{Over a field $k$, every subspace $U$ of a $k$-\textbf{vector space} $V$ is a direct summand}.'' Instructed four times to write module theory, the model relaxes into vector spaces.

This is the inheritance mechanism observed from the other side. A module-theoretic statement carrying an explicit field hypothesis is, in register terms, vector-space prose wearing module-theoretic nouns; the two are the same content at the same altitude, and the vocabulary that natively expresses that altitude exerts a pull the instruction does not overcome. The case also bounds an alternative reading of the identity controls: the drift is not random noise in framing compliance, since it occurs on exactly the chains where an altitude mismatch exists, and on no others.

\section{Legacy fidelity scale}
\label{app:legacy}

The $0$--$4$ ordinal with which fidelity was first scored is a deterministic function of the codes of Appendix~\ref{app:codebook}: $F=0$ if $T=\texttt{false}$; $4$ if $\texttt{equiv}\wedge\texttt{scope\_ok}$; $3$ if $\texttt{equiv}\wedge\neg\texttt{scope\_ok}$; $2$ if exactly one of $e_1,e_2$; $1$ otherwise. Every earlier table is therefore re-derivable. It was retired from inference for three reasons: on the cell that carries the paper the rubric forces $\bar F=2.5+k/18$ with $k$ the number of retained pins, so it is an affine re-encoding of pin retention, and reconstructing the earlier cells from their published means is unique and reproduces the published $F$-distribution exactly; it asserts an ordering between scope drift and one-way entailment that our data contradicts; and it places truth, an absolute property of the candidate, inside a relational scale. The composite $\mathrm{SPI}=(F/4)(C/3)$, where $C\in\{0,1,2,3\}$ is the legacy judged framing-compliance score superseded by \Cauto{}, and its gold-normalised variant are retained in the artefacts for backward compatibility only; the gold-normalised variant appears nowhere in this paper, being a ratio against a denominator that reached $0.25$ on one item.

\section{Supplementary tables}
\label{app:tables}

\paragraph{The control axes.} Table~\ref{tab:axes} gives truth, content equivalence, widening and
scope preservation by axis for the earlier two-model run, the only run in which the lateral and
functorial pairs were coded. It is the evidence for the claim in \S\ref{sec:setting} that the
control axes fail in a different way from the vertical one: content equivalence on the controls
falls as low as $48.9\%$ while widening stays near zero, whereas on the vertical axis content
equivalence is $92.6$--$97.8\%$ and widening reaches $41.5$--$64.4\%$. The controls lose content;
the vertical axis loses scope. Levels are not comparable with the main text, which uses the
repaired codebook.

\begin{table}[h]
\caption{\textbf{Failure by axis and direction} (earlier two-model run, original codebook, $k=3$).
A$\to$B on the vertical axis is abstracting. Percentages of codings.}
\label{tab:axes}
\centering
\footnotesize
\setlength{\tabcolsep}{4pt}
\begin{tabular}{@{}llrrrrr@{}}
\toprule
\textbf{Axis} & \textbf{Model} & \textbf{$n$} & \textbf{truth} & \textbf{content equiv.} & \textbf{widened} & \textbf{scope kept} \\
\midrule
vertical, abstracting & Claude Opus 5 & 135 & 94.8 & \textbf{92.6} & \textbf{64.4} & 46.7 \\
vertical, abstracting & GPT-5.6 Sol   & 135 & 98.5 & \textbf{97.8} & \textbf{41.5} & 63.7 \\
vertical, specialising & Claude Opus 5 & 135 & 100.0 & 95.6 & 0.0 & 77.0 \\
vertical, specialising & GPT-5.6 Sol   & 135 & 100.0 & 98.5 & 0.7 & 81.5 \\
\midrule
lateral, A$\to$B & Claude Opus 5 & 90 & 53.3 & \textbf{82.2} & 3.3 & 65.6 \\
lateral, A$\to$B & GPT-5.6 Sol   & 90 & 90.0 & \textbf{75.6} & 0.0 & 84.4 \\
functorial, A$\to$B & Claude Opus 5 & 45 & 93.3 & \textbf{48.9} & 2.2 & 68.9 \\
functorial, A$\to$B & GPT-5.6 Sol   & 45 & 64.4 & \textbf{84.4} & 8.9 & 88.9 \\
\bottomrule
\end{tabular}
\end{table}

\paragraph{Corpus composition.} The corpus is $90$ items, $15$ per framing pair, selected by a rule fixed before any statement was inspected and annotated under a witness audit. The \texttt{load} columns give the consequence of dropping each vertical item's pin: \emph{truth}, the statement becomes false (\texttt{lb\_truth}); \emph{ill}, it becomes ill-formed; \emph{ref}, it stays true but is asserted of a different class of objects; \emph{inert}, no effect. \emph{None} marks the seven items with no pin annotated, excluded from every pin-based denominator (hence $45-7-1=37$ in Table~\ref{tab:ladder}); \emph{n/a} marks the lateral and functorial items, which have no pin. The three released domains not measured here (ids $181$--$270$) are all lateral pairs, so their omission narrows the control axis and cannot affect a vertical result.

\begin{center}
\footnotesize
\setlength{\tabcolsep}{4pt}
\begin{tabular}{@{}lllrrrrrrr@{}}
\toprule
\textbf{axis} & \textbf{framing pair} & \textbf{ids} & \textbf{$n$} & \textbf{truth} & \textbf{ill} & \textbf{ref} & \textbf{inert} & \textbf{none} & \textbf{n/a} \\
\midrule
vertical & vector spaces $\leftrightarrow$ module theory & 1--29 odd & 15 & 2 & 1 & 1 & 11 & 0 & 0 \\
vertical & probability $\leftrightarrow$ measure theory & 31--59 odd & 15 & 6 & 3 & 1 & 1 & 4 & 0 \\
vertical & set theory $\leftrightarrow$ category theory & 61--89 odd & 15 & 1 & 8 & 0 & 3 & 3 & 0 \\
lateral & geometry $\leftrightarrow$ commutative algebra & 91--119 odd & 15 & 0 & 0 & 0 & 0 & 0 & 15 \\
lateral & graph theory $\leftrightarrow$ linear algebra & 151--179 odd & 15 & 0 & 0 & 0 & 0 & 0 & 15 \\
functorial & topology $\leftrightarrow$ algebra & 121--149 odd & 15 & 0 & 0 & 0 & 0 & 0 & 15 \\
\midrule
\multicolumn{10}{l}{\emph{released but not measured here:} discrete math $\leftrightarrow$ complex analysis (181--210), representation} \\
\multicolumn{10}{l}{theory $\leftrightarrow$ Fourier analysis (211--240), symmetric functions $\leftrightarrow$ tableaux (241--270)} \\
\bottomrule
\end{tabular}
\end{center}

\paragraph{Run sizes.}
\Mone{}: $966$ rewrites from $26$ design-blind contexts and $5{,}471$ codings from $43$ blind judging
contexts across three judge models. \Mtwo{}: a pre-registered replication with $540/540$ condition-A
rewrites (all $90$ items, both directions, $k=3$) and $54/54$ matched condition-\textbf{B} rewrites.
Vertical candidates were coded at $k=3$, lateral and functorial candidates at one sample per item and
direction; truth and repaired scope coding cover all $617$ canonical candidates (model outputs, MELD golds
and controls), a separately shuffled, generator-blind pool of $1{,}188$ candidates from $594$ byte-matched
tasks supplies the common-judge columns of Table~\ref{tab:perpair}, and $466$ failure-selected,
generator-blind candidates form the typology of Table~\ref{tab:typology}. A later \Mone{} run adds $846$
rewrites and $180$ blind codings for the instruction ladder and control axis (Appendix~\ref{app:ladder}).
The \Mtwo{} transfer, perturbation and round-trip runs were cut prospectively and are a declared boundary
of scope.

\paragraph{Three accounts of pin presence.} Each account predicts $P(\text{pin})$ for every framing
pair and direction: \emph{requirement} predicts $1$ when promoting and $0$ when absorbing; \emph{source
inheritance} predicts the pin rate of the source side; \emph{target-register imitation} predicts that of
the target side. The likelihood is binomial over the counts of Table~\ref{tab:dsi}, with
$p=\varepsilon+(1-2\varepsilon)\,\hat p$ and the lapse $\varepsilon$ fitted per account and model; a lapse
of $0.50$ means the account carries no information. Native pin rates over all $270$ released pairs stand
in for the corpus's own source rates. With three framing pairs the comparison is descriptive. Table~\ref{tab:accounts} reports differences in log-likelihood; the absolute values are:

\begin{center}
\footnotesize
\begin{tabular}{@{}lcccccc@{}}
\toprule
& \multicolumn{3}{c}{\textbf{\Mone{}}} & \multicolumn{3}{c}{\textbf{\Mtwo{}}} \\
\cmidrule(lr){2-4}\cmidrule(lr){5-7}
\textbf{account} & $\log L\uparrow$ & lapse$\downarrow$ & $\Delta\log L\uparrow$ & $\log L\uparrow$ & lapse$\downarrow$ & $\Delta\log L\uparrow$ \\
\midrule
requirement (theory morphism) & $-154.01$ & $0.50$ & $-21.23$ & $-150.26$ & $0.41$ & $-3.03$ \\
source inheritance & $\mathbf{-132.78}$ & $0.25$ & $0$ & $-153.91$ & $0.50$ & $-6.68$ \\
target-register imitation & $-153.96$ & $0.50$ & $-21.18$ & $\mathbf{-147.22}$ & $0.36$ & $0$ \\
\bottomrule
\end{tabular}
\end{center}

\paragraph{Coded relation between the candidate's domain and the source's} (condition \textbf{A}; the source
of the widening counts in \S\ref{sec:dissoc}). \Mone{} is coded under the original codebook by the
adjudicated majority of its three judges, \Mtwo{} under the repaired codebook; the \texttt{=} column is
therefore not comparable across models, the \texttt{sup} share of non-\texttt{=} codes is.

\begin{center}
\small
\setlength{\tabcolsep}{4pt}
\begin{tabular}{@{}llcccc@{\hskip 1.2em}cccc@{}}
\toprule
& & \multicolumn{4}{c}{\textbf{\Mone{}}} & \multicolumn{4}{c}{\textbf{\Mtwo{}}} \\
\cmidrule(lr){3-6}\cmidrule(lr){7-10}
\textbf{axis} & \textbf{direction} & \texttt{=} & \texttt{sup} & \texttt{sub} & \texttt{neq} & \texttt{=} & \texttt{sup} & \texttt{sub} & \texttt{neq} \\
\midrule
vertical & abstracting & 28 & \textbf{102} & 0 & 5 & 75 & \textbf{55} & 2 & 3 \\
vertical & specialising & 103 & 0 & 29 & 3 & 104 & 1 & 29 & 1 \\
lateral & A$\to$B & 65 & 1 & 0 & 24 & 24 & 0 & 4 & 2 \\
lateral & B$\to$A & 75 & 0 & 1 & 14 & 26 & 1 & 0 & 3 \\
functorial & A$\to$B & 33 & 0 & 0 & 12 & 12 & 3 & 0 & 0 \\
functorial & B$\to$A & 24 & 0 & 0 & 21 & 7 & 0 & 3 & 5 \\
\bottomrule
\end{tabular}
\end{center}

Of \Mtwo{}'s $55$ abstracting \texttt{sup} codes, $46$ carry no verified restriction span and are silent
widenings; \Mone{}'s $55$ abstracting candidates that fail \texttt{scope\_ok} are the $40.7\%$ of
\S\ref{sec:dissoc}.

\paragraph{The same table from the second model's initial, unrepaired judge pass.} All $95$
candidates of that single batch. The \texttt{scope} code discriminates; \texttt{flag} does not.

\begin{center}
\small
\begin{tabular}{lcccc}
\toprule
\textbf{\texttt{scope} code} & \texttt{=} & \texttt{sup} & \texttt{sub} & \texttt{neq} \\
\midrule
count & 68 & 7 & 13 & 7 \\
of which \texttt{flag}$=1$ & 0 (by definition) & \textbf{7/7} & \textbf{13/13} & \textbf{7/7} \\
\bottomrule
\end{tabular}
\end{center}

\paragraph{Notation and transport.}\label{app:vocab}
\Mtwo{}'s mechanical figures come from an independently written implementation of the same specification,
so agreement between the models is an instrument replication as well as a model comparison; both main runs
contain $n=540$ rewrites. With every multi-letter word stripped, no pair falls below the exclusion
threshold of $0.85$ on either model, under two independently written tokenizers, and the pre-registered
minimum pairwise AUC of $0.90$ for \Mtwo{} is met at $0.921$ (Table~\ref{tab:mech}). The transport figures
of Table~\ref{tab:mech} are computed, as the benchmark's own analysis could not be, on the models' rewrites
\citep{ye2026doesembeddingreflecta}: for \Mtwo{} under the word representation \textbf{W}, rank-$1$
retrieval of the cross-framing twin rises from $26.1\%$ to $45.7\%$; the surviving \Mone{} baseline records
one pooled transport figure, $30.0\%$ to $48.0\%$, which we compare with \Mtwo{}'s \textbf{W} value. Under
notation alone (\textbf{N}), computed for \Mtwo{} only, retrieval rises from $14.4\%$ to $30.7\%$. That
$14.4\%$ is the one pre-registered acceptance check that failed, the notation representation having been required to place source-side rank-$1$ retrieval in a $20$--$35\%$ band; it is reported rather than tuned away, it belongs to a different representation from the $30.0\%$ above and does not contradict it.
Under \textbf{W}, \textbf{N} and \textbf{X}, on every pair and for both models, \Cauto{} orders a verbatim
copy below the model's rewrite and the rewrite below the held-out gold.

\paragraph{Sample-to-sample instability with the judge held fixed} (\Mone{}, abstracting, vertical).

\begin{center}
\small
\begin{tabular}{lccc}
\toprule
\textbf{framing pair} & $k=1$ & $k=2$ & $k=3$ \\
\midrule
vector spaces $\to$ module theory & 14/15 & 8/15 & 11/15 \\
probability $\to$ measure theory & 15/15 & 12/15 & 10/15 \\
set theory $\to$ category theory & 10/15 & 0/15 & 1/15 \\
\midrule
\textbf{pooled} & \textbf{39/45} & \textbf{20/45} & \textbf{22/45} \\
\bottomrule
\end{tabular}
\end{center}

\paragraph{Inter-rater reliability of the superseded first run.} These figures come from the three same-family judges of the original run, not from the instrument used in \S\ref{sec:exp}; the primary code of that instrument transfers between judge families at $\kappa=1.000$ and agrees with a blind expert reading at $\kappa=0.900$ (\S\ref{sec:instrument}). Krippendorff's $\alpha$, computed on items with $\ge2$ raters: \texttt{equivalent} (PP binary) $\mathbf{0.807}$ ($n=210$); \texttt{scope} (4-way) $0.599$; $e_2$ $0.538$; \texttt{truth} $0.478$; $e_1$ $0.467$; \texttt{flag} $0.436$ (all $n=914$ except truth, $n=250$). Simplicity buys reliability, which is the argument for a judge-free compliance axis.

\section{Statistical procedure, pre-registration and decision rules}
\label{app:prereg}

\paragraph{Statistical procedure.} The $k$ samples of an item are not exchangeable and rewrites within an item are not independent, so the unit of analysis throughout is the item. For each item we take the mean of its $k$ codings in each direction and test the paired difference with an exact sign-flip permutation test, enumerating all $2^{n}$ sign assignments where $n \le 20$ and sampling $5\times10^{4}$ otherwise. Confidence intervals are cluster bootstraps over items, stratified by framing pair where a pooled estimate is reported. Contingency comparisons use Fisher's exact test. Because one arm of the direction contrast is near zero, the paired test is arithmetically close to a one-sample test of the abstracting rate; we therefore report that rate with a binomial interval alongside every contrast rather than leaving the equivalence implicit.


\section{The regex--judge convergence on three populations}
\label{app:convergence}
The same quantity, namely the rate at which a candidate that states the standing hypothesis is judged to have widened the domain against the rate for one that omits it, computed on each of the three populations in which it can be formed. \textbf{A}: Claude Opus 5 alone, both halves,
$65$ candidates, $7.1\%$ against $90.2\%$. \textbf{B}: seven models, held-out items, $315$
candidates, $2.1\%$ against $85.9\%$ (the value quoted in \S\ref{sec:instrument}). \textbf{C}:
three models on the ProofWiki items, $72$ scorable candidates, where only $3$ state the hypothesis at all
and the comparison is uninformative.

\section{Additional experiments on Claude and GPT}
\label{app:superseded}
This appendix reports a second set of experiments on two models, Claude Opus 5 and GPT-5.6 Sol, at
$k=3$ over all $90$ items and all three axes. It covers analyses that the seven-model run does not:
the lateral and functorial control axes, the per-pair typology, and the comparison of accounts of
hypothesis presence. Its judged codes were produced under an earlier version of the scope codebook
rather than the repaired one used in the main text, so its levels are not directly comparable with
those of \S\ref{sec:exp} and are read here for their internal contrasts.
\subsection*{Results}
\label{old:sec:results}

These experiments proceed from phenomenon to mechanism to remedy. They locate the loss of scope
(\S\ref{old:sec:dissoc}), characterise the failures by type and cost (\S\ref{sec:typology}) and,
because both analyses rest on judged codes, use the judge-free measure to identify what governs
hypothesis presence (\S\ref{sec:dsi}) and test whether instruction restores directional control
(\S\ref{sec:ladder}). Uncertainties are one standard error and tests are exact and
two-sided. The corresponding analyses in the present revision are reported in \S\ref{sec:exp}
under the repaired codebook, and the levels below are not comparable with them.

\paragraph{Content is preserved; generality is not}
\label{old:sec:dissoc}

\begin{table}[t]
\caption{\textbf{Truth, content and scope under the bare prompt}: percentage of rewrites coded true,
content-equivalent on the shared domain, and scope-preserving (\texttt{scope\_ok}).
\Mone{} is coded by the adjudicated majority of three judges and \Mtwo{} by one judge under the repaired
codebook (\S\ref{sec:judgefail}); vertical cells pool $k=3$ samples, and \Mtwo{}'s control cells use one.
$n$: \Mone{}\,/\,\Mtwo{}.}
\label{tab:main}
\begin{center}
\small
\setlength{\tabcolsep}{4pt}
\begin{tabular}{@{}llrcccccc@{}}
\toprule
& & & \multicolumn{3}{c}{\textbf{\Mone{} (Claude Opus 5)}} & \multicolumn{3}{c}{\textbf{\Mtwo{} (GPT-5.6 Sol)}} \\
\cmidrule(lr){4-6}\cmidrule(lr){7-9}
\textbf{axis} & \textbf{direction} & $n$ & truth$\uparrow$ & equiv.$\uparrow$ & \textbf{scope}$\uparrow$ & truth$\uparrow$ & equiv.$\uparrow$ & \textbf{scope}$\uparrow$ \\
\midrule
vertical & abstracting (promote) & 135\,/\,135 & 98.5 & 99.3 & \textbf{59.3} & 98.5 & 98.5 & \textbf{63.7} \\
vertical & specialising (absorb) & 135\,/\,135 & 100 & 97.8 & 95.6 & 100 & 98.5 & 81.5 \\
\midrule
lateral & A$\to$B & 90\,/\,30 & 94.4 & 95.6 & 95.6 & 90.0 & 66.7 & 90.0 \\
lateral & B$\to$A & 90\,/\,30 & 95.6 & 95.6 & 100 & 93.3 & 76.7 & 93.3 \\
functorial & A$\to$B & 45\,/\,15 & 100 & 95.6 & 88.9 & 66.7 & 86.7 & 80.0 \\
functorial & B$\to$A & 45\,/\,15 & 100 & 86.7 & 86.7 & 93.3 & 53.3 & 80.0 \\
\bottomrule
\end{tabular}
\end{center}
\end{table}

If a model represents the content of a statement but not its level of generality, truth and content
should survive translation, while scope should fail where a hypothesis must be introduced, namely when
abstracting on vertical pairs, and Table~\ref{tab:main} confirms this prediction. On vertical pairs, truth and
content equivalence lie between $97.8$ and $100\%$ in every cell, whereas scope is preserved in
$59.3\pm4.2\%$ of \Mone{}'s abstracting rewrites and $63.7\pm4.1\%$ of \Mtwo{}'s (Fisher exact test, $p=0.53$), against $95.6\%$ and $81.5\%$ when specialising. The controls isolate
the effect: with no generality to lose, scope is preserved in $80.0$--$100\%$ of rewrites, so vertical
abstraction is the only cell below $80\%$; the controls lose content instead (\Mtwo{}'s equivalence falls to
$53.3\%$), which an equivalence check detects. The vertical losses are silent: $40.7\%$ of \Mone{}'s and
$36.3\%$ of \Mtwo{}'s abstracting rewrites change the domain without stating it, almost always by widening
($102$ of $107$ and $55$ of $60$ non-identity domain codes are \texttt{sup}; Appendix~\ref{app:tables}), and
on the shared domain such a rewrite is identical to its source.

\begin{table}[t]
\caption{\textbf{Abstracting scope preservation by framing pair} (\%). \emph{Canonical}: each
model's registered coding, with a Fisher exact test. \emph{Common judge}: both models' rewrites of
byte-identical tasks re-coded blind to generator by one judge. \emph{MELD gold}: reference translations ($n=30$ per pair) under \Mtwo{}'s judge.}
\label{tab:perpair}
\begin{center}
\small
\setlength{\tabcolsep}{5pt}
\begin{tabular}{@{}lccc@{\hskip 1.4em}cc@{\hskip 1.4em}c@{}}
\toprule
& \multicolumn{3}{c}{\textbf{canonical coding}} & \multicolumn{2}{c}{\textbf{common judge}} & \textbf{reference} \\
\cmidrule(lr){2-4}\cmidrule(lr){5-6}\cmidrule(lr){7-7}
\textbf{framing pair, abstracting} & \Mone{}$\uparrow$ & \Mtwo{}$\uparrow$ & $p$ & \Mone{}$\uparrow$ & \Mtwo{}$\uparrow$ & MELD gold$\uparrow$ \\
\midrule
vector spaces $\to$ modules & 73.3 & 33.3 & $3{\times}10^{-4}$ & 33.3 & 33.3 & 96.7 \\
probability $\to$ measure & 80.0 & 64.4 & $0.16$ & 66.7 & 66.7 & 43.3 \\
set theory $\to$ categories & \textbf{24.4} & \textbf{93.3} & $1.2{\times}10^{-11}$ & \textbf{40.0} & \textbf{91.1} & 83.3 \\
\midrule
pooled & 59.3 & 63.7 & $0.53$ & & & 74.4 \\
\bottomrule
\end{tabular}
\end{center}
\end{table}

Since pooled rates may conceal heterogeneity and judge effects, Table~\ref{tab:perpair}
disaggregates them. Under canonical coding the endpoints reverse: \Mone{} preserves
scope best on vector spaces and worst on set theory ($73.3$ against $24.4\%$), and \Mtwo{} shows the
opposite pattern ($33.3$ against $93.3\%$). Re-coded by one judge, the models coincide on two pairs and
differ only on set theory $\to$ categories ($40.0$ against $91.1\%$), so part of the canonical difference
reflects the judge, which motivates the judge-free analysis of \S\ref{sec:dsi}. The benchmark's reference
translations preserve scope on $96.7\%$ of vector-space pairs but only $43.3\%$ of probability pairs, where
gold-referenced scoring could not detect the relocation.

\paragraph{Failures relocate rather than falsify}
\label{sec:typology}

We next characterise the failures by type, which reflects the operation performed, and by cost, which
determines whether outcome-based evaluation could detect them. On vertical pairs, failures are coded as a
generalised conclusion (\dC{}) and a coerced type (\dT{}), with ``vector space'' read as ``module'', and
rarely as an omitted hypothesis (\dH{}: $3.1\%$ on vector spaces $\leftrightarrow$ modules;
Table~\ref{tab:typology}), as expected when the pin never appears in the source text. The remaining types are confined to the pairs whose structure admits them:
referent drift occurs only on geometry $\leftrightarrow$ commutative algebra ($22$ of $22$ codes), the one
anti-equivalence, where a dictionary applied covariantly yields the dual object (Appendix~\ref{app:cases}),
and illocutionary drift mainly on graph theory $\leftrightarrow$ linear algebra ($9$ of $11$).

\begin{wraptable}{r}{0.53\textwidth}
\vspace{-1.1\baselineskip}
\centering
\caption{\textbf{Abstracting scope preservation by pin load.} Each model is scored under its own,
independently produced \texttt{load} annotation (\Mone{}'s has no ill-formed category); $p$: Fisher exact
test, truth against the rest.}
\label{tab:load-old}
\footnotesize
\setlength{\tabcolsep}{2.5pt}
\begin{tabular}{@{}lcc@{}}
\toprule
\textbf{dropping the pin makes it\dots} & \textbf{\Mone{}}$\uparrow$ & \textbf{\Mtwo{}}$\uparrow$ \\
\midrule
false (\emph{truth}) & \textbf{21/24} (87.5) & \textbf{26/27} (96.3) \\
ill-formed & n/a & 31/36 (86.1) \\
true of other objects (\emph{ref.}) & 12/30 (40.0) & \phantom{0}2/6\phantom{0} (33.3) \\
unchanged (\emph{inert}, none) & 47/81 (58.0) & 27/66 (40.9) \\
\midrule
$p$, truth vs.\ rest & $0.0024$ & $2.9{\times}10^{-5}$ \\
\bottomrule
\end{tabular}
\vspace{-0.6\baselineskip}
\end{wraptable}
With respect to cost (Table~\ref{tab:load-old}), both models preserve scope where omitting the pin would make
the statement false ($87.5\%$ and $96.3\%$), and \Mtwo{} where it would make it ill-formed ($86.1\%$), but
not where the widened statement remains true ($40.0\%$ and $33.3\%$ under reference load; $58.0\%$ and
$40.9\%$ under inert load). The loss is thus concentrated where no truth-checker, prover or equivalence
score would register it, subject to three qualifications: the item-level crosswalk between the two
annotations was not retained, ill-formed load is concentrated on set theory (eight of twelve items), and
\Mtwo{}'s reference cell rests on six rewrites. The cause of these failures is examined next with the judge-free measure.

\paragraph{Generality is inherited rather than derived}
\label{sec:dsi}

A model that derived generality from the requirements of the translation would state the pin when promoting
and omit it when absorbing ($\DSI\approx+100$); one that copied the source's surface would yield a negative
\DSI{}. Table~\ref{tab:dsi} measures both directions without a judge. \textbf{\Mone{}'s \DSI{} is
negative}: pooled, it states the pin in $18.0\pm3.6\%$ of rewrites where it is required and $47.7\pm4.7\%$
where it is redundant. On vector spaces and modules, it states the field hypothesis in $82.2\%$ of rewrites
\emph{into vector-space prose}, where the noun encodes it, and $4.4\%$ \emph{into module-theory prose},
where it is required, a nineteen-fold difference in the wrong direction; paired within item, $5$ items carry
the pin only when abstracting against $16$ only when specialising (exact McNemar test,
$p=0.027$). \Mtwo{}'s pooled \DSI{} is positive ($+18.0$; $41.4\pm4.7\%$ against $23.4\pm4.0\%$) but again
describes no single pair: it controls the pin on set theory $\leftrightarrow$ categories ($+66.7$) alone.

\begin{table}[t]
\caption{\textbf{Pin presence in both directions} (judge-free, $k=3$): percentage of rewrites
containing the registered pin. Promotion requires the pin ($100\%$ is correct), absorption its removal ($0\%$
is correct; a retained pin is redundant, not false); \DSI{} in points. The item whose abstracting source
contains the pin is excluded. Right: pin frequency in native prose over all $270$ MELD pairs.}
\label{tab:dsi}
\begin{center}
\footnotesize
\setlength{\tabcolsep}{3.5pt}
\begin{tabular}{lccc@{\hskip 0.9em}ccc@{\hskip 0.9em}cc}
\toprule
& \multicolumn{3}{c}{\textbf{\Mone{} Claude Opus 5}} & \multicolumn{3}{c}{\textbf{\Mtwo{} GPT-5.6 Sol}} & \multicolumn{2}{c}{\textbf{native prose}} \\
\cmidrule(lr){2-4}\cmidrule(lr){5-7}\cmidrule(lr){8-9}
\textbf{framing pair} & promote$\uparrow$ & absorb$\downarrow$ & \DSI$\uparrow$ & promote$\uparrow$ & absorb$\downarrow$ & \DSI$\uparrow$ & general & concrete \\
\midrule
vector spaces $\leftrightarrow$ modules & 4.4\% & 82.2\% & $\mathbf{-77.8}$ & 17.8\% & 28.9\% & $-11.1$ & 30/30 & \textbf{0/30} \\
probability $\leftrightarrow$ measure & 20.0\% & 40.0\% & $-20.0$ & 30.0\% & 26.7\% & $+3.3$ & 10/30 & \textbf{1/30} \\
set theory $\leftrightarrow$ categories & 33.3\% & 11.1\% & $+22.2$ & 80.6\% & 13.9\% & $+66.7$ & 25/30 & \textbf{0/30} \\
\midrule
\textbf{pooled} & \textbf{18.0\%} & \textbf{47.7\%} & $\mathbf{-29.7}$ & \textbf{41.4\%} & \textbf{23.4\%} & $\mathbf{+18.0}$ & & \\
\bottomrule
\end{tabular}
\end{center}
\end{table}

\begin{table}[t]
\begin{minipage}[t]{0.48\textwidth}
\caption{\textbf{Three accounts of pin presence} (Table~\ref{tab:dsi} counts): log-likelihood
relative to each model's best account and fitted lapse $\varepsilon$ ($0.50$: uninformative).}
\label{tab:accounts}
\centering
\footnotesize
\setlength{\tabcolsep}{2.5pt}
\begin{tabular}{@{}lcccc@{}}
\toprule
& \multicolumn{2}{c}{\textbf{\Mone{}}} & \multicolumn{2}{c}{\textbf{\Mtwo{}}} \\
\cmidrule(lr){2-3}\cmidrule(lr){4-5}
\textbf{account} & $\Delta\log L\uparrow$ & $\varepsilon\downarrow$ & $\Delta\log L\uparrow$ & $\varepsilon\downarrow$ \\
\midrule
requirement & $-21.2$ & $0.50$ & $-3.0$ & $0.41$ \\
source inheritance & $\mathbf{0}$ & $0.25$ & $-6.7$ & $0.50$ \\
target register & $-21.2$ & $0.50$ & $\mathbf{0}$ & $0.36$ \\
\bottomrule
\end{tabular}
\end{minipage}\hfill
\begin{minipage}[t]{0.48\textwidth}
\caption{\textbf{Judge-free mechanical axes.} AUC: leave-one-out separation of a pair's two
framings (range over pairs). Twin: retrieval of, and cosine to, the cross-framing twin
(Appendix~\ref{app:vocab}).}
\label{tab:mech}
\centering
\footnotesize
\setlength{\tabcolsep}{3pt}
\begin{tabular}{@{}lcc@{}}
\toprule
& \textbf{\Mone{}} & \textbf{\Mtwo{}} \\
\midrule
AUC, words (\textbf{W}) & .894--1 & .951--1 \\
AUC, notation (\textbf{N}) & \textbf{.909--1} & \textbf{.921--.998} \\
AUC, char.\ $n$-grams (\textbf{X}) & .902--1 & .962--1 \\
twin rank-1 (\%)$\,\uparrow$ & 30.0$\,\to\,$48.0 & 26.1$\,\to\,$45.7 \\
twin cosine$\,\uparrow$ & .149$\,\to\,$.348 & .145$\,\to\,$.333 \\
\bottomrule
\end{tabular}
\end{minipage}
\end{table}

\paragraph{Three accounts of pin presence.}
The divergent per-pair profiles suggest different policies. We fit three accounts by binomial likelihood with one lapse rate (Table~\ref{tab:accounts}; Appendix~\ref{app:tables}):
\emph{requirement}, under which the pin is always stated when promoting and never when absorbing;
\emph{source inheritance}, under which the source's pin state is copied; and \emph{target-register
imitation}, under which the pin is stated as often as native prose in the target dialect states it. Source
inheritance describes \Mone{} decisively, leading both alternatives by $21.2$ in log-likelihood, while both fit only at chance ($\varepsilon=0.50$). Target-register imitation best describes \Mtwo{},
ahead of requirement by $3.0$ and of inheritance by $6.7$, a modest, descriptive margin over three pairs; consistently, its one controlled pair is the one whose pin, ``In \Set'', is native
category-theoretic vocabulary, whereas module theory opens with ``let $R$ be a ring'', which the pin must
override. For neither model is the requirement the best account.

\paragraph{Surface recoverability of framings.}
Target-register imitation presupposes that a framing's register is recoverable from its surface form, as
Table~\ref{tab:mech} confirms: with every multi-letter word removed, the two framings of a pair separate at
leave-one-out AUC of $0.909$--$1.000$ (\Mone{}) and $0.921$--$0.998$ (\Mtwo{}), above
a pre-registered $0.90$. Translation moves a statement toward its cross-framing twin but leaves it more than
half short in both models (Appendix~\ref{app:vocab}).

\end{document}